%% file: main.tex
\documentclass[twoside]{article}
\usepackage[accepted]{aistats2024}

\usepackage{amsmath}
\usepackage{amsthm}
\usepackage{algorithm, algorithmic}

\usepackage{natbib}

\usepackage{amssymb}
\usepackage{url}
\usepackage{enumitem}
\usepackage{sidecap}

\usepackage{stmaryrd}
\usepackage{textcomp}

\usepackage{subfigure}
\usepackage[utf8]{inputenc} 
\usepackage[T1]{fontenc}    
\usepackage{hyperref}       
\usepackage{amsfonts}       
\usepackage{nicefrac}       
\usepackage{microtype}      

\usepackage[usenames, dvipsnames]{color}

\usepackage{thm-restate}

\newtheorem{theorem}{Theorem}[section]

\newtheorem{lemma}[theorem]{Lemma}

\newtheorem{definition}[theorem]{Definition}

\newcommand{\nc}[1]{\newcommand{#1}}
\nc{\bs}[1]{\boldsymbol{#1}}

\usepackage{stmaryrd}
\usepackage{textcomp}

\usepackage{tikz,lipsum,lmodern}
\usepackage[most]{tcolorbox}

\nc{\nid}{K}
\nc{\idc}[1]{c_{#1}}
\nc{\idv}{\bs{c}}
\nc{\iddv}{\bs{D}}
\nc{\idd}[2]{D_{#1,#2}}
\nc{\emp}[1]{\emph{#1}}
\nc{\ntr}{T}
\nc{\al}{\circ}
\nc{\dal}{\bullet}
\nc{\cho}[1]{\mathcal{A}_{#1}}
\nc{\indi}[1]{\llbracket#1\rrbracket}
\nc{\mis}{L}
\nc{\be}{\begin{equation*}}
\nc{\ee}{\end{equation*}}
\nc{\siz}{M}
\nc{\nat}{\mathbb{N}}
\nc{\hsiz}{S}
\nc{\pvt}[1]{\bs{p}_{#1}}
\nc{\pct}[2]{p_{#1,#2}}
\nc{\splx}[1]{\Delta_{#1}}
\nc{\pset}[1]{\mathcal{P}_{#1}}
\nc{\qset}[1]{\mathcal{Q}_{#1}}
\nc{\lr}{\eta}
\nc{\hal}[2]{\ell_{#1,#2}}
\nc{\hav}[1]{\bs{\ell}_{#1}}
\nc{\tpv}[1]{\tilde{\bs{p}}_{#1}}
\nc{\tpc}[2]{\tilde{p}_{#1,#2}}
\nc{\aws}[1]{\mathcal{D}_{#1}}
\nc{\acv}[1]{\bs{a}_{#1}}
\nc{\acc}[2]{a_{#1,#2}}
\nc{\st}[1]{s_{#1}}
\nc{\sid}{i^*}
\nc{\mt}[1]{\mu_{#1}}
\nc{\ret}[1]{r_{#1}}
\nc{\ars}{\mathcal{S}}
\nc{\bexpt}[1]{\mathbb{E}\left[#1\right]}
\nc{\prob}[1]{\mathbb{P}[#1]}

\nc{\af}{r}
\nc{\did}{\tilde{r}}
\nc{\bas}{\mathcal{Q}}
\nc{\nlm}[2]{\Lambda^{#1}(#2)}
\nc{\nps}[2]{\Phi^{#1}(#2)}
\nc{\acq}[2]{b_{#1}(#2)}
\nc{\asq}[1]{\mathcal{B}(#1)}
\nc{\tmp}{x}
\nc{\npf}[1]{\Phi^{#1}}
\nc{\mat}[2]{\bs{U}^{#1,#2}}
\nc{\mac}[4]{U^{#1,#2}_{#3,#4}}
\nc{\nep}[2]{\epsilon^{#1,#2}}
\nc{\tra}[1]{#1^\top}
\nc{\ave}{\bs{x}}
\nc{\avec}[1]{x_{#1}}
\nc{\qc}[1]{\mathcal{C}_{#1}}
\nc{\acqp}[2]{b'_{#1}(#2)}
\nc{\asqp}[1]{\mathcal{B}'(#1)}
\nc{\dir}[1]{\hat{r}_{#1}}
\nc{\apr}{\alpha}
\nc{\tpo}{s}
\nc{\lgt}{\log_2}
\nc{\rct}{c}
\nc{\ses}{K}
\nc{\cov}[1]{\mathcal{C}_{#1}}
\nc{\bet}[1]{\beta_{#1}}
\nc{\mco}{\mathcal{S}}
\nc{\cco}{\gamma}
\nc{\sca}{\varepsilon}
\nc{\bprob}[1]{\mathbb{P}\left[#1\right]}
\nc{\prp}{s}
\nc{\mmu}{c}
\nc{\kex}{d}
\nc{\raf}{g}
\nc{\haf}{h}
\nc{\ssi}{B}
\nc{\kld}[2]{\delta(#1,#2)}
\nc{\lay}{f}
\nc{\lyt}[1]{f_{#1}}
\nc{\nma}{K}
\nc{\lye}{\tilde{f}}
\nc{\aac}{\mathcal{A}}
\nc{\ass}{\mathcal{S}}
\nc{\lfs}{\mathcal{F}}
\nc{\lft}[1]{f_{#1}}
\nc{\ctv}[1]{\bs{c}_{#1}}
\nc{\ctc}[2]{c_{#1,#2}}
\nc{\arm}{a}
\nc{\prof}{P}
\nc{\egr}[2]{g_{#1,#2}}
\nc{\gra}[1]{\bs{g}_{#1}}
\nc{\nmb}{K}
\nc{\prot}[1]{\psi_{#1}}
\nc{\cms}{\mathcal{X}}
\nc{\reg}[1]{R(#1)}
\nc{\obf}[1]{\Psi_{#1}}
\nc{\psti}[2]{p^{#1}_{#2}}
\nc{\pstv}[1]{\bs{p}^{#1}}
\nc{\red}[1]{\tilde{r}_{#1}}
\nc{\asre}{z}
\nc{\mxc}{C}
\nc{\fac}[1]{\mathcal{A}'_{#1}}
\nc{\exs}[1]{\Gamma_{t,#1}}
\nc{\ext}[2]{\tilde{\Gamma}_{t,#1,#2}}
\nc{\apv}{\bs{q}}
\nc{\apc}[1]{q_{#1}}
\nc{\drt}[1]{\tilde{r}_{#1}}
\nc{\exr}[1]{\Upsilon_{#1}}
\nc{\caj}[2]{\Lambda_{#1,#2}}
\nc{\ca}[1]{\Lambda_{#1}}
\nc{\ter}[1]{\Psi_{#1}}
\nc{\ex}[1]{\Gamma_{#1}}
\nc{\rplus}{\mathbb{R}_+}
\nc{\br}{\mathbb{R}}
\nc{\cex}[2]{\expt{#1\,|\,#2}}
\nc{\cpr}[2]{\prob{#1\,|\,#2}}
\nc{\cbex}[2]{\bexpt{#1\,\biggr|\,#2}}
\nc{\rgt}{R}
\nc{\pro}[1]{\psi_{#1}}
\nc{\pstc}[1]{p^*_{#1}}
\nc{\fun}[1]{f_{#1}}
\nc{\gpv}{\bs{q}}
\nc{\gpc}[1]{q_{#1}}
\nc{\der}{\nabla}
\nc{\pder}[1]{\partial_{#1}}
\nc{\pos}{j'}
\nc{\fth}[3]{\kappa_{#1,#2,#3}}
\nc{\sth}[2]{\lambda_{#1,#2}}
\nc{\expt}[1]{\mathbb{E}[#1]}
\nc{\pst}{\bs{p}^*}
\nc{\psc}[1]{p^*_{#1}}
\nc{\sun}[1]{\sigma_{#1}}
\nc{\prl}[2]{\hat{\ell}_{#1,#2}}
\nc{\alg}{\textsc{MSE3}}
\nc{\nmu}{N}
\nc{\tco}[1]{\gamma_{#1}}
\nc{\nml}{L}
\nc{\flalg}{FLE3}
\nc{\phr}[1]{\Phi^{#1}}
\nc{\asr}{\mathcal{Z}}
\nc{\asrt}{\mathcal{Z}'}
\nc{\cas}{\sigma}
\nc{\ltm}{m}
\nc{\fpi}{z}
\nc{\asm}[1]{\mathcal{Z}_{#1}}
\nc{\dif}{\phi}
\nc{\lz}[1]{z_{#1}}
\nc{\nni}{\mu}
\nc{\sgs}{\sigma^*}

\renewcommand{\tilde}{\widetilde}

\begin{document}

\twocolumn[
\aistatstitle{Sum-max Submodular Bandits}


\aistatsauthor{ Stephen Pasteris \And Alberto Rumi \And  Fabio Vitale \And Nicol\`{o} Cesa-Bianchi}

\aistatsaddress{ Alan Turing Institute, \\ United Kingdom \And  Universit\`{a} degli Studi di Milano \\ and CENTAI, Italy \And CENTAI, Italy \And Università degli Studi di Milano \\ and Politecnico di Milano, Italy } ]
\begin{abstract}
Many online decision-making problems correspond to maximizing a sequence of submodular functions. In this work, we introduce sum-max functions, a subclass of monotone submodular functions capturing several interesting problems, including best-of-$K$-bandits, combinatorial bandits, and the bandit versions on facility location, $M$-medians, and hitting sets. We show that all functions in this class satisfy a key property that we call pseudo-concavity. This allows us to prove $\big(1 - \frac{1}{e}\big)$-regret bounds for bandit feedback in the nonstochastic setting of the order of $\sqrt{MKT}$ (ignoring log factors), where $T$ is the time horizon and $M$ is a cardinality constraint. This bound, attained by a simple and efficient algorithm, significantly improves on the $\widetilde{\mathcal{O}}\big(T^{2/3}\big)$ regret bound for online monotone submodular maximization with bandit feedback.
\end{abstract}

\input{intro-ncb}

\input{sum-max}

\input{related}

\section{MAIN RESULT} \label{sec:problemdefinition}

Our learning problem is formally defined as follows. The values $\siz,\nmb\in\nat$ and $\mxc\in\br_+$ are all preliminarily known to the learner. Hidden from the learner, the adversary selects a sequence of 
set functions $\langle\ret{t}\,|\,t\in[\ntr]\rangle$, each with domain $2^{[\nmb]}$ and a sequence of vectors $\langle\ctv{t}\,|\,t\in[\ntr]\rangle$ each in $[0,\mxc]^{\nmb}$. On each trial $t\in[\ntr]$:
\begin{enumerate}[nosep]
\item The learner chooses some $\cho{t}\subseteq[\nmb]$ with $|\cho{t}|\leq\siz$.
\item The value $\ret{t}(\cho{t})$ is revealed.
\item For all $i\in\cho{t}$ the value $\ctc{t}{i}$ is also revealed.
\end{enumerate}

The learner maintains a probability vector $\pvt{t}\in\splx{\nmb}$, and behaves as described in Algorithm \ref{alg:msc}.

\begin{algorithm}[t] 
\caption{\alg}\label{alg:msc}
Set $\lr:=\ln(\nmb)/\rgt$ and $\pct{1}{i}:=1/\nmb$ for $i\in[\nmb]$

\smallskip
\textbf{for} $t=1,2,\ldots, T$ \textbf{do}:
\begin{enumerate}[wide,nosep]
\item For all $j\in[\siz]$ draw $\acc{t}{j}\in[\nmb]$ from 
distribution $\pvt{t}$
\item Define $\cho{t}:=\{\acc{t}{j}~|~j\in[\siz]\}$
\item Receive $\ret{t}(\cho{t})$ and $\{\ctc{t}{i}\,|\,i\in\cho{t}\}$
\item For all $i\in[\nmb]$ set
\be
\egr{t}{i}:=\frac{\ret{t}(\cho{t})-\ctc{t}{i}}{\pct{t}{i}}\sum_{j\in[\siz]}\indi{\acc{t}{j}=i}
\ee
\item For all $i\in[\nmb]$ define $\tpc{t}{i}:=\pct{t}{i}\exp(\lr\egr{t}{i})$
\item Define $\pvt{t+1}:=\tpv{t}/\|\tpv{t}\|_1$
\end{enumerate}
\end{algorithm}

To aid our theorem statement we add the following definitions.
For all $t\in[\ntr]$ and $\bas\subseteq[\nmb]$ we define $\dir{t}(\bas):=\ret{t}(\bas)-\ret{t}(\emptyset)$, which is the difference between the learner's profit on trial $t$ and that which it would have obtained by selecting the empty set, $\pro{t}:=\dir{t}(\cho{t})-\tco{t}(\cho{t})$, and
$
\tco{t}(\bas):=\sum_{i\in\bas}\ctc{t}{i}
$.
We note that by considering $\dir{t}$ instead of $\ret{t}$ our bounds do not change when $\ret{t}$ is shifted by an additive constant (which can be different for different trials $t$) as long as the range of $\ret{t}$ falls within the bounds described as follows.

We assume that the learner knows upper and lower bounds on the range $\ret{t}$ for all trials $t$. Hence, without loss of generality, assume that $\ret{t}(\bas)\in[-1,0]$ for all $t\in[\ntr]$ and $\bas\setminus[\nmb]$ (otherwise scale and shift $\ret{t}$ and $\mxc$).
Let 
\be
\rgt:=(1+\mxc)\sqrt{2\ln(\nmb)\siz(\nmb+\siz-1)\ntr}\,.
\ee
Our results hold for a relaxed notion of submodularity, which we call \emp{pseudo-submodularity}.

\begin{definition}
    A set function $\af:2^{[\nmb]}\rightarrow\br$ is \emp{pseudo-submodular} if and only if for every set $\ass\subseteq[\nmb]$ with $\ass\neq\emptyset$ there exists some $i\in\ass$ such that for all $\bas\subseteq\ass\setminus\{i\}$ we have
$
\af(\bas\cup\{i\})-\af(\bas)\geq\af(\ass)-\af(\ass\setminus\{i\})
$.
\end{definition} Note that all pseudo-submodular set functions are also submodular. We now present our main result.
\begin{restatable}{theorem}{mainth}\label{mainth}
Given $\ret{t}$ is pseudo-concave and pseudo-submodular for all $t\in[T]$\,, then for any set $\ass\subseteq[\nmb]$ with $\ass\neq\emptyset$ we have
\be
\sum_{t\in[\ntr]}\expt{\pro{t}}\geq\left(1-\alpha^\siz\right)\sum_{t\in[\ntr]}\dir{t}(\ass)-\frac{\siz}{|\ass|}\sum_{t\in[\ntr]}\tco{t}(\ass)-\rgt
\ee
where
$
\alpha := 1-\frac{1}{|\ass|}
$.
\end{restatable}

\begin{proof}
See Section \ref{analysec}
\end{proof}

We note that both the standard facility location and $k$-medians problems are often phrased as the minimization of a loss rather than a maximization of a profit. Our results easily capture this by considering the reward as a negative loss.
%

We now show that the approximation ratio $1-\alpha^{\siz}$ is not improvable in general in the class of sum-max functions. In particular, we show that obtaining an efficient online learning algorithm for the multichannel advertising problem with a sublinear $\gamma$-regret with $\gamma < 1-\alpha^{\siz}$ would give an efficient randomized algorithm for solving set cover on $[\nmb]$ with an approximation better than $\ln\nmb$. As shown in \citep{dinur2014analytical}, obtaining an approximation of $(1-\varepsilon)\ln\nmb$ for set cover is NP-hard for any $\varepsilon > 0$.

Recall that an instance of the multichannel campaign problem over $\nmb$ ads is defined by a sequence $\langle \ret{t}\,|\,t\in[\ntr]\rangle$ of set functions over $[\nmb]$ such that for all $t\in[T]$ there exists some $\aws{t}\subseteq[\nmb]$ with
$
\ret{t}(\bas)=\indi{\bas\cap\aws{t}\neq\emptyset}
$
for all $\bas\subseteq[\nmb]$.
\begin{restatable}{theorem}{thLB}\label{Th:LB1}
Suppose that there exists some $\kex\in\nat$, $\tpo\in(0,1)$, $\cco > 1$, and a randomized polynomial time algorithm for the learner such that for all $\nmb, \siz\in\nat$ and for any instance of the multichannel advertising problem, it holds that $\big|\cho{t}\big|\le \siz$ for all $t=1,\ldots,T$ and, for any subset $\mco\subseteq[\nid]$,
\be
\bexpt{\sum_{t\in[\ntr]}\ret{t}(\cho{t})}\geq\left(1-\alpha^{\cco\siz}\right)\sum_{t\in[\ntr]}\ret{t}(\ars)- R'\,,
\ee
where $R' \in \mathcal{O}(\nid^\kex\ntr^\tpo)$ and
$
\alpha := 1 - \frac{1}{|\ars|}
$.
Then, for all $\sca \in \big(0,1-1/\cco\big)$ and $\ssi>4^{1/((1-\sca)\cco-1)}$, there exists a randomized polynomial-time algorithm for the set cover problem on $[\ssi]$ that, with probability at least $\frac{1}{2}$, achieves approximation ratio at least $(1-\sca)\ln(\ssi)$.
\end{restatable}

The proof can be found in Appendix \ref{apx:LB1}

\section{BANDIT FACILITY LOCATION}


\renewcommand{\nmb}{L}
\renewcommand{\nml}{K}
\begin{algorithm}[t]
\caption{\flalg}\label{alg:flalg}
Run \alg\ with $\nmb = 2\nml$ arms and
$
\siz:=\frac{\nml}{2}\ln(\ntr/\nml^2)
$.

\smallskip
On each trial $t\in[\ntr]$:
\begin{enumerate}[nosep]
\item Let $\cho{t}'$ be the output of \alg
\item Output $\cho{t}:=\cho{t}'\cap[\nml]$
\item Receive $\ret{t}(\cho{t})$ and $\{\ctc{t}{i}\,|\,i\in[\nml]\}$
\item For all $i\in[\nmb]\setminus[\nml]$ set $\ctc{t}{i}:=0$
\item Feed $\ret{t}(\cho{t})$ and $\{\ctc{t}{i}\,|\,i\in[\nmb]\}$ back to \alg
\end{enumerate}
\end{algorithm}

In this application, there are no restrictions on the set of arms $\cho{}$ that we choose. We seek to maximize $\af(\cho{})-\gamma(\cho{})$ where $\af$ is the sum-max reward function and $\gamma$ is the linear and positive cost function.


For the facility location problem we must choose $\siz$, noting that although a high value of $\siz$ increases the approximation ratio on the reward, it also increases that on the costs. To decrease the potentially large approximation ratio on the costs, we borrow from \cite{Pasteris2020OnlineLO} the idea of \emph{dummy arms} and the tuning of $\siz$. This leads to our algorithm \flalg\ described in Algorithm \ref{alg:flalg}. The bound on the total profit of \flalg\ is given in the following theorem.

\begin{restatable}{theorem}{fleth}\label{fle3th}
Given that 
$C=1$ and $\ret{t}:2^{\nml}\rightarrow[-1,0]$ is pseudo-concave and pseudo-submodular for all $t\in[\ntr]$\,, we have that the algorithm $\flalg$ obtains the following bound for all $\ass\subseteq[\nml]$ with $\ass\neq\emptyset$:
\be
\sum_{t\in[\ntr]}\expt{\pro{t}}\geq\sum_{t\in[\ntr]}\dir{t}(\ass)-\frac{1}{2}\ln\left(\frac{\ntr}{\nml^2}\right)\sum_{t\in[\ntr]}\tco{t}(\ass)-R''\,,
\ee
where
$
R''\in \tilde{\mathcal{O}}(\nml\sqrt{\ntr})
$.
\end{restatable}



\begin{proof}
For all $t\in[\ntr]$ define the set function $r'_t:2^{\nmb}\rightarrow[0,1]$ such that for all $\bas\subseteq[\nmb]$, $r'_t(\bas):=\ret{t}(\bas\cap[\nml])$ and, as consequence, $ \hat{r}'_t(\bas):=r'_t(\bas)-r'_t(\emptyset)$. 
Now taking into consideration any possible comparator set $\mathcal{S} \subseteq [K]$, we define
\be
\ass':=\ass\cup\{\nml+i\,|\,i\in[\nml-|\ass|]\}\,,
\ee
noting that $|\ass'|=\nml$.
Note that $r'_t$ is sum-max and hence, by Theorem \ref{summaxth}, is pseudo-concave and submodular for all $t\in[\ntr]$. This allows us to apply Theorem \ref{mainth}, that gives us:
\begin{align}
\nonumber
    &\sum_{t\in[\ntr]}\expt{\pro{t}}\geq\left(1-\alpha^\siz\right)\sum_{t\in[\ntr]}\hat{r}'_{t}(\ass')-\frac{\siz}{|\ass'|}\sum_{t\in[\ntr]}\tco{t}(\ass')-\rgt
\\&=
\label{eq:rprime}
    \left(1-\alpha^\siz\right)\sum_{t\in[\ntr]}\hat{r}_{t}(\ass)-\frac{\siz}{|\ass'|}\sum_{t\in[\ntr]}\tco{t}(\ass)-\rgt
\\&=
\label{eq:def}
\left(1-\alpha^\siz\right)\sum_{t\in[\ntr]}\hat{r}_{t}(\ass)-\frac{1}{2}\ln\frac{\ntr}{\nml^2}\sum_{t\in[\ntr]}\tco{t}(\ass)-\rgt\,,
\end{align}
where equation (\ref{eq:rprime}) comes from the contribution of the dummy arms and equation (\ref{eq:def}) from the definition of $M$. Given that
\begin{align*}
\alpha &:= \frac{|\ass'|-1}{|\ass'|}=\frac{\nml-1}{\nml}\leq \exp\left(-{1}/{\nml}\right),
\end{align*}
we can therefore see that
\begin{align}
\nonumber
    \alpha^{\siz}\sum_{t\in[\ntr]}\hat{r}_{t}(\ass)
&\leq
    \alpha^{\siz}\ntr
=
    \exp(-\siz/\nml)\ntr
\\&=
    \frac{1}{\sqrt{\ntr/\nml^2}}\ntr
=
    \sqrt{\ntr\nml^2}\,,
\label{eq:alpha}
\end{align}
where we used the definition of $\siz$ given in Algorithm~\ref{alg:flalg}.
Putting together (\ref{eq:def}) and (\ref{eq:alpha}) gives us the result, where $R'' = R + \sqrt{\ntr\nml^2}$.
\end{proof}

\renewcommand{\nmb}{K}
\renewcommand{\nml}{L}

\section{ANALYSIS}\label{analysec}

We now give an overview of the proof of Theorem \ref{mainth}.

We first consider the case that we have no costs (i.e. $\ctv{t}=\bs{0}$). \alg\ works by maintaining a probability distribution over the set of arms. Specifically, $\pvt{t}\in\splx{\nmb}$ is the vector whose components are the probabilities of drawing the actions on trial $t$. On trial $t$ the algorithm constructs the set $\cho{t}$ by drawing a sequence $\langle\acc{t}{j}\,|\,j\in[\siz]\rangle$ of arms i.i.d. with replacement from $\pvt{t}$ and then setting $\cho{t}:=\{\acc{t}{j}\,|\,j\in[\siz]\}$.

This stochastic draw of a sequence and set from a probability vector will be represented by the following notation.

\begin{definition}
For all $\apv\in\splx{\nmb}$ let $\langle\acq{j}{\apv}\,|\,j\in[\siz]\rangle$ be a sequence of stochastic quantities drawn i.i.d. at random from (the probability distribution characterised by) $\apv$. In addition, let
$
\asq{\apv}:=\{\acq{j}{\apv}\,|\,j\in[\siz]\}
$.
\end{definition}

Note that our expected reward on trial $t$ is $\expt{\ret{t}(\asq{\pvt{t}})}$ and hence, for all set functions $\af$ we shall construct a differentiable function $\phr{\af}:\mathbb{R}^{\nmb}\rightarrow\mathbb{R}$ such that for all $\apv\in\splx{\nmb}$ we have $\phr{\af}(\apv)=\expt{\af(\asq{\apv})}$. This construction is based on the following notion of a \emph{subset decomposition}.

\begin{definition}
Given a function $\af:2^{[\nmb]}\rightarrow\br$\,, we call a function $\did:2^{[\nmb]}\rightarrow\br$ a \emp{subset decomposition} of $\af$ if and only if for all $\bas\subseteq[\nmb]$ we have
\be
\af(\bas)=\sum_{\ass\subseteq[\nmb]}\indi{\bas\subseteq\ass}\did(\ass)\,.
\ee
\end{definition}

The following lemma confirms that every set function has a unique subset decomposition.

\begin{restatable}{lemma}{lemmasubsetcomp}\label{lemma:subsetcomp}
Given a function $\af:2^{[\nmb]}\rightarrow\br$ there exists a
unique subset decomposition $\did$ of $\af$.
\end{restatable}

\begin{proof}
See Appendix \ref{apx:subsetcomp}.
\end{proof}

Now we can define our function $\phr{\af}$.

\begin{definition}
For all $\af:2^{\nmb}\rightarrow\br$ and all $\apv\in\br^{\nmb}$ define
\be
\nps{\af}{\apv}:=\sum_{\ass\subseteq[\nmb]}\did(\ass)\left(\sum_{i\in[\nmb]}\indi{i\in\ass}\apc{i}\right)^\siz\,,
\ee
where, by Lemma \ref{lemma:subsetcomp}, $\did$ is the unique subset decomposition of $\af$.
\end{definition} 

The following lemma confirms that our function $\phr{\af}$ indeed satisfies our condition.

\begin{restatable}{lemma}{pheqexrelem}\label{pheqexrelem}
For all $\af:2^{\nmb}\rightarrow\br$ and all $\apv\in\splx{\nmb}$ we have
$
\nps{\af}{\apv}=\expt{\af(\asq{\apv})}
$.
\end{restatable}

\begin{proof}
See Appendix \ref{apx:pheqexrelem}
\end{proof}

Drawing inspiration from \cite{Auer2001TheNM} we will learn via online exponentiated gradient ascent with the functions $\phr{\ret{t}}$ using unbiased gradient estimates. Of course, this means that we must be able to construct unbiased gradient estimates. Remarkably, we now show that we can use our sequence $\langle\acc{t}{j}\,|\,j\in[\siz]\rangle$ and the observed reward $\ret{t}(\cho{t})$ to construct an unbiased gradient estimate $\gra{t}$ defined in Algorithm \ref{alg:msc} of the function $\phr{\ret{t}}$ at $\pvt{t}$.

\begin{restatable}{lemma}{derphilem}\label{derphilem1}
For all $\af:2^{\nmb}\rightarrow\br$, all $\apv\in\splx{\nmb}$ and all $i\in[\nmb]$ we have
\be
\pder{i}\nps{\af}{\apv}=\bexpt{\frac{\af(\asq{\apv})}{\apc{i}}\sum_{j\in[\siz]}\indi{\acq{j}{\apv}=i}}\,.
\ee
\end{restatable}

\begin{proof}
See Appendix \ref{apx:derphilem1}
\end{proof}

For exponentiated gradient ascent to work, we must have that, for all trials $t$, our objective function $\phr{\ret{t}}$ is concave over the simplex. We now show that a sufficient condition for this to hold is that the function $\ret{t}$ is pseudo-concave.

\begin{restatable}{lemma}{phiconlem}\label{phiconlem1}
For all pseudo-concave set functions $\af:2^{\nmb}\rightarrow\br$ we have that $\npf{\af}$ is concave over the simplex $\splx{\nmb}$.
\end{restatable}

\begin{proof}
See Appendix \ref{apx:phiconlem1}
\end{proof}


Now that we have all the underpinnings for exponentiated gradient ascent to function properly, we can establish a bound on the regret relative to any vector $\pst\in\splx{\nmb}$ via the following classic result.

\begin{lemma}\label{hedgelem1}
For any vector $\pst\in\splx{\nmb}$ we have
\begin{align*}
\sum_{t\in[\ntr]}(\pst-\pvt{t})\cdot\gra{t}\leq~&\frac{1}{\lr}\sum_{i\in[\nmb]}\pstc{i}\ln(\nmb\pstc{i})\\
&+\lr\sum_{t\in[\ntr]}\sum_{i\in[\nmb]}\pct{t}{i}\egr{t}{i}^2\,.
\end{align*}
\end{lemma}

\begin{proof}
A classic result from the analysis of \textsc{Hedge}.
\end{proof}

This lemma gives a bound on the regret since, because we have shown that $\gra{t}$ is an unbiased estimate of the gradient and the objective function is concave over the simplex, the term $(\pst-\pvt{t})\cdot\gra{t}$ is bounded below by $\phr{\ret{t}}(\pst)-\phr{\ret{t}}(\pvt{t})$. Note that we have shown above that $\phr{\ret{t}}(\pvt{t})$ is equal to $\expt{\ret{t}(\cho{t})}$.

We will later discuss the bounding of the regret itself, but first we shall show how to choose $\pst$ such that we can bound $\phr{\ret{t}}(\pst)$ relative to $\ret{t}(\ass)$ for some set $\ass\subseteq[\nmb]$. Specifically, we will choose $\pst$ equal to $\pstv{\ass}$ in the following definition.

\begin{definition}
For all $\ass\subseteq[\nmb]$ with $\ass\neq\emptyset$ define $\pstv{\ass}\in\splx{\nmb}$ such that for all $i\in[\nmb]$ we have
\be
\psti{\ass}{i}:=\frac{\indi{i\in\ass}}{|\ass|}\,.
\ee
\end{definition}

We use the following lemma will to bound $\phr{\ret{t}}(\pstv{\ass})$, and it explains why we require $\ret{t}$ to be pseudo-submodular.

\begin{restatable}{lemma}{psubrlem}\label{psubrlem1}
Let $\ass\subseteq[\nmb]$ with $\ass\neq\emptyset$, $\af:2^{[\nmb]}\rightarrow\br$ be a pseudo-submodular function, and $\asr\subseteq[\nmb]$ be a set formed by drawing $\siz$ elements uniformly at random (with replacement) from $\ass$. Then we have
\be
\expt{\af(\asr)-\af(\emptyset)}\geq\left(1-\left(\frac{|\ass|-1}{|\ass|}\right)^\siz\right)(\af(\ass)-\af(\emptyset))\,.
\ee
\end{restatable}

\begin{proof}
See Appendix \ref{apx:psubrlem2}
\end{proof}

With this lemma in hand, we can now bound $\phr{\ret{t}}(\pstv{\ass})$.

\begin{restatable}{lemma}{expslem}\label{expslem}
Given any $\ass\subseteq[\nmb]$ and any pseudo-submodular set function $\af:2^{[\nmb]}\rightarrow\br$ we have
\be
\nps{\af}{\pstv{\ass}}\geq\af(\emptyset)+\left(1-\left(\frac{|\ass|-1}{|\ass|}\right)^\siz\right)(\af(\ass)-\af(\emptyset))
\ee
\end{restatable}

\begin{proof}
See Appendix \ref{apx:expslem}.
\end{proof}

Before we bound the regret term, we show how to incorporate the costs, so that $\ctv{t}$ can be non-zero. This is done by choosing, instead of $\phr{\ret{t}}$, the objective function $\obf{t}$ defined as follows.

\begin{definition}
For all trials $t\in[\ntr]$ define $\obf{t}:\br^{\nmb}\rightarrow\br$ such that for all $\apv\in\br^{\nmb}$ we have
\be
\obf{t}(\apv):=\npf{\ret{t}}(\apv)-\siz\apv\cdot\ctv{t}\,.
\ee
\end{definition}

Note that by Lemma \ref{pheqexrelem} we have that  $\obf{t}(\pvt{t})$ is a lower bound on the expected profit and by Lemma \ref{phiconlem1} $\obf{t}$ is concave over the simplex. It can hence serve as a surrogate concave objective function.

Lemma \ref{derphilem1} leads to the following lemma, which confirms that $\gra{t}$ is an unbiased gradient estimate of $\obf{t}$ at $\pvt{t}$.

\begin{restatable}{lemma}{unbgreslem}\label{unbgreslem1}
For all trials $t\in[\ntr]$ we have
\be
\nabla\obf{t}(\pvt{t})=\cex{\gra{t}}{\pvt{t}}
\ee
\end{restatable}

\begin{proof}
See Appendix \ref{apx:unbgreslem1}, \end{proof}

Now we have shown that our results carry over to the case of non-zero costs, we can finally bound the regret via Lemma \ref{hedgelem1} and the following lemma.

\begin{figure*}[t]
  \centering

  \subfigure[Stochastic]{
    \includegraphics[width=0.3\hsize]{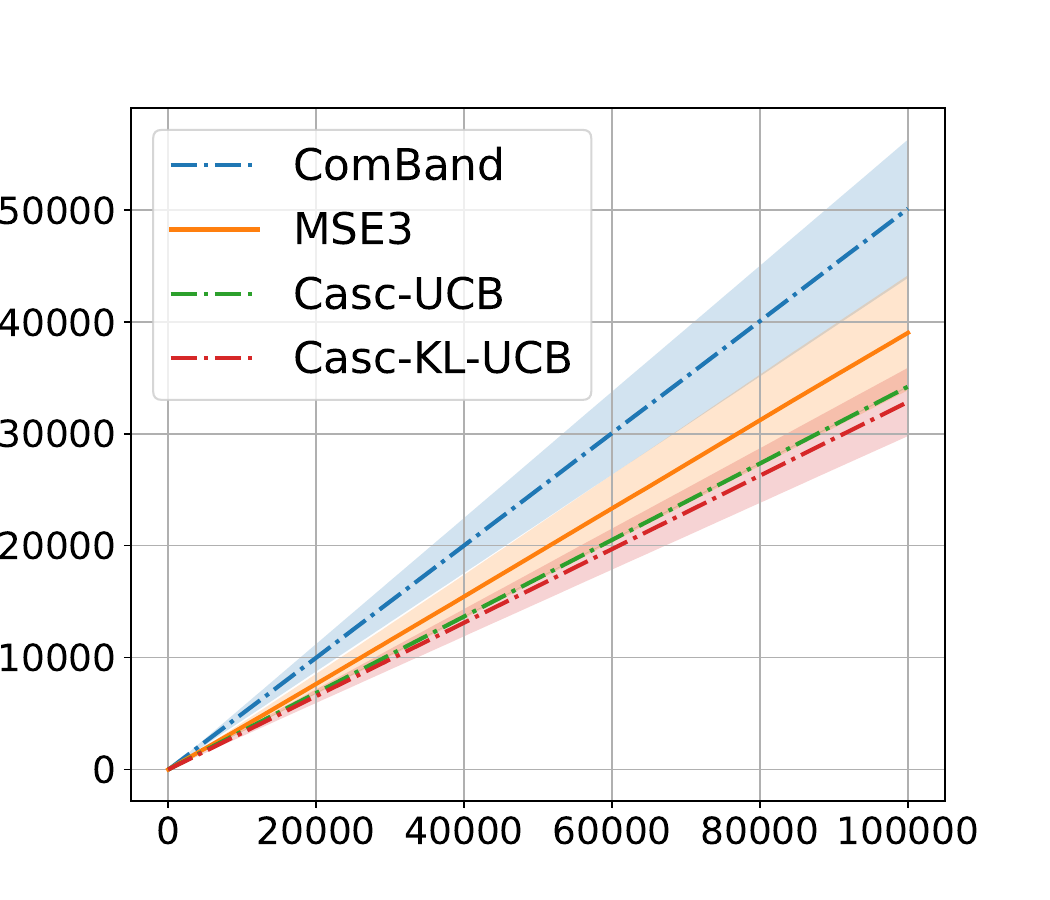}
    \label{fig:stochastic}
  }
  \hfill
  \subfigure[Stochastic with adversarial corruptions]{
    \includegraphics[width=0.3\hsize]{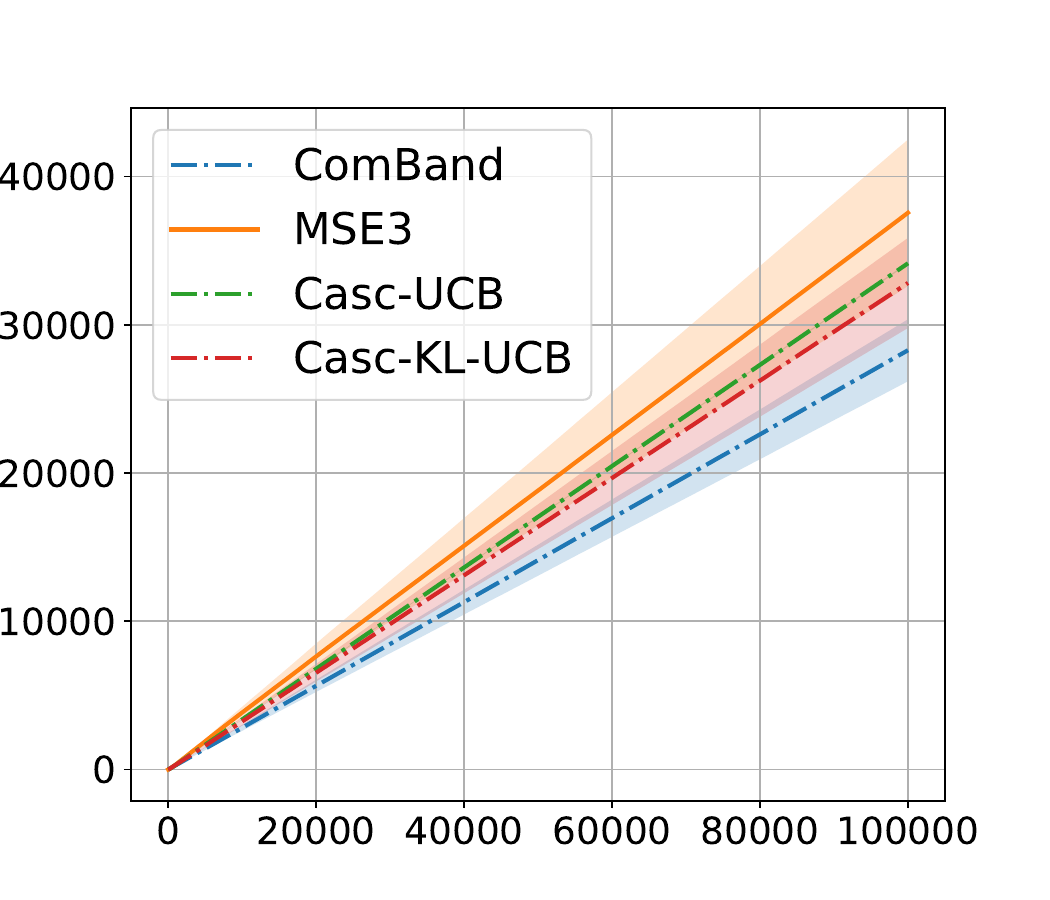}
    \label{fig:corrupted}
  }
  \subfigure[Worst-case stochastic]{
    \includegraphics[width=0.3\hsize]{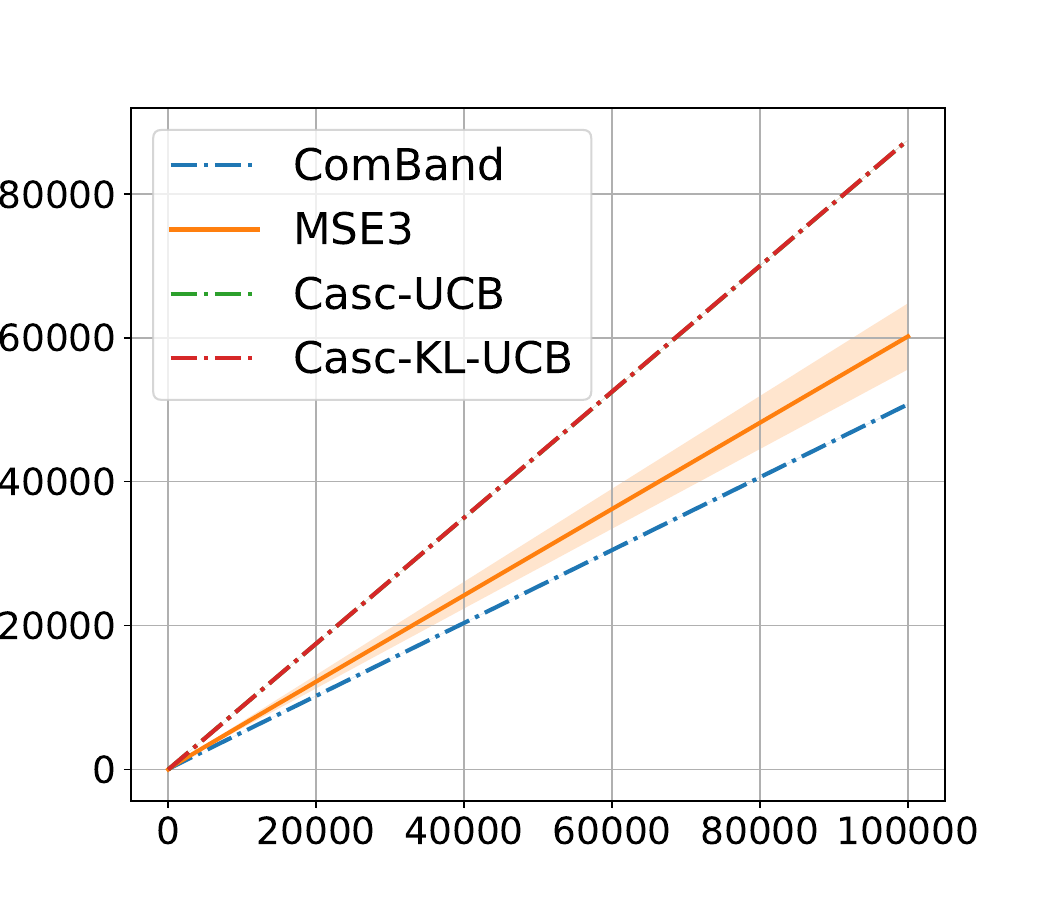}
    \label{fig:lowerbound}
  }
  \caption{Cumulative reward over time in the three environment settings is described. The results also display the $95\%$ confidence intervals over $35$ runs with an Intel Xeon Gold 6312U, calculated using the standard error multiplied by the $z$-score of $1.96$.}
  \label{fig:mainfigure}
\end{figure*}

\begin{restatable}{lemma}{susqlem}\label{susqlem}
For all trials $t\in[\ntr]$ we have
\be
\bexpt{\sum_{i\in[\nmb]}\pct{t}{i}\egr{t}{i}^2}\leq(1+\mxc)^2\siz(\nmb+\siz-1)\,.
\ee
\end{restatable}

\begin{proof} 
See Appendix \ref{apx:susqlem}.
\end{proof}

This completes the analysis. Although discussed here,  Appendix \ref{apx:final_lemma} formally shows how to piece the lemmas together in order to prove Theorem \ref{mainth}.

\section{EXPERIMENTS}

We experimentally evaluated the performance of our method by comparing it with two baselines: \textsc{CascadeBandit} from \cite{kveton2015cascading} (in both the \textsc{UCB} and \textsc{KL} settings) and \textsc{Comband} from \cite{cesa2012combinatorial} for $M$-sized subsets, whose efficient implementation is described in Appendix \ref{apx:comband}. We conducted our experiments in various synthetic settings. In each of these environments, a hidden vector $\boldsymbol{\theta} \in \mathbb{R}^K$ is maintained. For each $k \in [K]$, the entry $\theta_k$ represents the probability of obtaining a unit reward. 
These values can be viewed as \textit{attraction probabilities}: the probability that a user clicks on the specific item. After presenting a subset of $M$ elements, the learner gets a unit reward if any of the selected items returns a 1, and 0 otherwise. It is worth emphasizing that our model does not necessitate binary rewards; it offers the flexibility to accommodate any sum-max reward function (as discussed in Section \ref{sec:sum-max}). The use of a binary reward model is specifically required for comparisons with click models as \textsc{CascadeBandit}.

\paragraph{Environments for the experiments.} We experimentally evaluated our method in three different synthetic environments. We conducted experiments across a wide range of values for $K$, $M$, $T$, and for the probabilities associated with both optimal and suboptimal arms. In Figure~\ref{fig:mainfigure}, we display the cumulative reward over time obtained with $T = 10^5$, $K = 20$, $M = 3$ when the environments are set as follows:

\begin{enumerate}[wide]
    \item \textbf{Stochastic} (Figure~\ref{fig:stochastic}): we randomly select $M$ \textit{good actions} to which we assign a reward probability of $0.3$. The reward probabilities of the remaining $k-M$ arms are set to $0.1$.
    
    \item \textbf{Stochastic with adversarial corruptions} (Figure~\ref{fig:corrupted}): the rewards are generated as in the stochastic setting. However, in the first $\sqrt{T}$ rounds all good actions have a deterministic reward of $0$.

    \item \textbf{Worst-case stochastic} (Figure~\ref{fig:lowerbound}): this setting is inspired by the lower bound of \citet{cohen2017tight}. Here the set $\mathcal{M} \subset [K]$ of $M$ \textit{good actions} is drawn uniformly at random. Then, for each $k \in [K]$, the probabilities are assigned as follows:
    \begin{align*}
    \theta_k &= 
    \begin{cases}
    X_k + \epsilon & \text{if } k \in \mathcal{M}
    \\
    X_k & \text{otherwise}
    \end{cases}
\quad\text{where}\quad
    X_k \sim N\left(\frac{1}{2}, \sigma^2\right)
\\
    \sigma^2 &= \frac{1}{192 + 96\log{T}}
    \quad\text{and}\quad
    \epsilon=\sigma\sqrt{\frac{KM}{8T}}
    \end{align*}

\end{enumerate}

\paragraph{Results} 

As expected, our most compelling results were achieved in the adversarial setting, where our approach demonstrated its superiority. In the two stochastic settings, we observed results that were on par with the established baseline methods, affirming the competitiveness of our proposed approach. 
These findings collectively underscore the effectiveness of our method, particularly in the challenging adversarial context, while also highlighting its versatility in stochastic scenarios. We emphasize that our method is the most efficient one, as each prediction only requires sampling $M$ times from a probability distribution over the $K$ available actions.


\section{FUTURE WORK}
In this work we gave a $(1-1/e)$-regret bound of $\mathcal{O}(\sqrt{T})$ for a specific class of functions which intersects with monotone submodular set-functions. Can we achieve such a bound for all monotone submodular set-functions? A crucial property used in this work is that of pseudo-concavity. Can we characterize other classes of submodular pseudo-concave functions? For instance, are all budget-additive functions pseudo-concave? Since the standard adversarial bandit problem is a special case of our problem, we know that a regret of $\Omega(\sqrt{KT})$ is required. Can we prove that a regret of $\Omega(\sqrt{MKT})$ is required?

\bibliography{bibliography}
\bibliographystyle{abbrvnat}

\input{supplemental}

\end{document}

%% file: intro-ncb.tex
\section{INTRODUCTION}
In many concrete settings of sequential decision-making, decisions are subsets of a finite set $[\nmb]$ (possibly with cardinality constraints) and utilities, or rewards, are non-linear set functions over $[\nmb]$. Although we may know that utility functions have some specific structure, e.g., they are submodular, the feedback may not reveal anything beyond the utility of the current decision. For example, consider an advertising campaign over $[\nmb]$ digital channels (e.g., web, apps, and social media). Due to budget constraints, the campaign can show ads only on a subset of $\siz$ channels for every user. If a user ends up buying the advertised product, we observe that a sale occurred, but we may not know which of the $\siz$ channels triggered the purchase. The advertiser's goal is to choose the subset of channels for each new user in order to maximize the number of sales.

The same problem was studied (with a different motivation) by \citet{simchowitz2016best} under stochastic assumptions on the generation of the Bernoulli random variables each indicating whether displaying an ad on a certain channel triggers a purchase for the current user. In this work, we study the nonstochastic variant of this problem, where the binary variables associated with the channels are chosen, for each user, by an oblivious adversary. Our main result is an efficient algorithm minimizing regret in a much larger class of problems containing the multichannel advertising problem as a special case. In particular, our regret analysis applies to any sequential decision-making problem where reward functions belong to a subclass of all monotone submodular functions called \emph{sum-max}.

A sum-max function is defined by a nonnegative matrix with $\nmb$ columns and an arbitrary number of rows. The value of the function evaluated at a subset $\ass\subset [\nmb]$ of columns is the sum over the rows of the maximum row element over the subset $\ass$ of columns. In the multichannel campaign example, the matrix is binary with a single row. The $j$-th entry indicates whether the current user would buy the product if advertised on channel $j$. If the matrix is square and symmetric, then we recover the non-metric facility location problem as a special case. 



As we said earlier, our analysis of regret for sum-max functions assumes bandit feedback: at each time $t$ we only observe the reward $\ret{t}(\cho{t})$ associated with our decision $\cho{t}$, where $\ret{t}$ is the sum-max function chosen by the adversary at time $t$. Hence, the reward $\ret{t}(\ass)$ that we would have obtained by choosing any $\ass\neq\cho{t}$ remains unknown.  We also consider cardinality constraints, in the form of a parameter $\siz$ requiring that the decision $\cho{t}$ at each time $t$ satisfy $\big|\cho{t}\big| \le \siz$. Note that when $\siz=1$ we recover the adversarial $\nmb$-armed bandit problem.

Our main result is an efficient algorithm, \alg, achieving a $\widetilde{\mathcal{O}}\big(\sqrt{\siz\nmb\ntr}\big)$ bound on the $\gamma_{\siz}$-regret for $\gamma_{\siz} = 1 - \big(1-1/\siz\big)^{\siz}$.
For comparison, for the class of all monotone submodular functions, \citet{niazadeh2021online} obtain a $\big(1-\frac{1}{e}\big)$-regret bound of $\mathcal{O}\big((\ln\nmb)^{1/3}\siz(\nmb\ntr)^{2/3}\big)$. As $\gamma_{\siz} > 1-\frac{1}{e}$ for all $\siz > 1$, this bound is worse than ours in both approximation factor and regret.

When $\siz=1$, algorithm \alg\ reduces to the standard \textsc{Exp3} algorithm for $\nmb$-armed bandits and our result specializes to the standard $\mathcal{O}\big(\sqrt{\nmb(\ln\nmb)\ntr}\big)$ regret bound of \textsc{Exp3}. This implies that the $\sqrt{\nmb\ntr}$ dependence in the regret bound is not improvable, even disregarding efficiency.
Moreover, we show that improving on the approximation factor $\gamma_{\siz}$ with an efficient algorithm would give an efficient randomized algorithm for solving set cover on $[\nmb]$ with an approximation ratio of $(1-\varepsilon)\ln\nmb$, which is NP-hard for any $\varepsilon > 0$ \citep{dinur2014analytical}.

In many real world problems, including an element $i$ in the decision $\cho{t}$ at round $t$ invokes a cost (i.e., a negative reward) $\ctc{t}{i} \ge 0$. When this is the case we would like to maximize the cumulative reward:
\be
\sum_{t\in[\ntr]}\left(\ret{t}(\cho{t})-\sum_{i\in\cho{t}}\ctc{t}{i}\right)
\ee
We show that \alg\ can handle this generalised problem if it receives, at the end of each round $t$, the values of $\ctc{t}{i}$ for all $i\in\cho{t}$. We note that the bandit \alg\ without costs is a special case of \alg\ with costs.

The inclusion of costs creates a tension between including arms in $\cho{t}$ to increase the reward and, simultaneously, avoid including too many arms to control the costs. We address this trade-off by introducing and analysing a variant of \alg\ for regret minimization with costs and bandit feedback where the the rewards are sum-max functions without cardinality constraints. We call this setting the bandit facility location problem because it is a bandit version of the online facility location problem studied by \citet{Pasteris2020OnlineLO}. 

For $\siz > 1$ and arbitrary costs, \alg\ selects $\cho{t}$ by performing $\siz$ independent draws $\acc{t}{1},\ldots,\acc{t}{\siz}$ from a distribution $\pvt{t} = \big(\pct{t}{1},\ldots,\pct{t}{\nmb}\big)\in\splx{\nmb}$. Then, a reward estimate for each $i \in [\nmb]$ is computed using
\begin{equation}
\label{eq:estimate}
    \egr{t}{i}=\frac{\ret{t}(\cho{t})-\ctc{t}{i}}{\pct{t}{i}}\sum_{j\in[\siz]}\indi{\acc{t}{j}=i}\,,
\end{equation}
where, for any statement \( S \), the Iverson bracket notation \( \indi{\cdot} \) is defined as \( \indi{S} = 1 \) if \( S \) is true and \( \indi{S} = 0 \) otherwise. Note that for $\siz=1$ and $\ctc{t}{i}=0$\,, the above reduces to the standard reward estimate of \textsc{Exp3}.

We now give an overview of how \alg\ works when we have no costs (i.e., $\ctc{t}{i}=0$).
For all set functions $\af$, we construct a function $\phr{\af}:\mathbb{R}_+^{\nmb}\rightarrow\mathbb{R}$ such that for all $\apv\in\splx{\nmb}$ we have that $\phr{\af}(\apv)$ is the expected value of $\af(\cho{})$ when $\cho{}$ is constructed by drawing $\siz$ arms i.i.d.\ with replacement from $\apv$. Specifically, we first show that there exists a function $\did:2^{[\nmb]}\rightarrow\mathbb{R}$ such that for all $\bas\subseteq[\nmb]$ we have $\af(\bas)=\sum_{\ass\subseteq[\nmb]}\indi{\bas\subseteq\ass}\did(\ass)$. For all $\apv\in\mathbb{R}_+^{\nmb}$ we then define: 
\be
\phr{\af}(\apv)=\sum_{\ass\subseteq[\nmb]}\did(\ass)\left(\sum_{i\in[\nmb]}\indi{i\in\ass}\apc{i}\right)^\siz\,.
\ee

We learn via online exponentiated gradient ascent using the unbiased estimates \eqref{eq:estimate} of the gradient of $\phr{\ret{t}}$.
Clearly, for exponentiated gradient ascent to work we must have that, for all rounds $t$, our objective function $\phr{\ret{t}}$ is concave over the simplex. We show that a sufficient condition for this to hold is that the function $\ret{t}$ is pseudo-concave, see Section~\ref{sec:sum-max} for a formal definition.

Next, we bound the regret with respect to any vector $\pst\in\splx{\nmb}$. Namely, we bound the expected reward of our algorithm relative to $\sum_{t\in[T]}\phr{\ret{t}}(\pst)$. By taking $\pst$ such that $p^*_i=\indi{i\in\ass}/|\ass|$ for some set $\ass$ we show that, because $\ret{t}$ is submodular, we have $\phr{\ret{t}}(\ass)\geq(1-\alpha^\siz)\ret{t}(\ass)$ where $\alpha=(|\ass|-1)/|\ass|$. By bounding the variance of the gradient estimate we show that the regret term is $\widetilde{\mathcal{O}}(\sqrt{\siz\nmb\ntr})$.

We have provided an overview of how, when we have no costs, \alg\ works and why we require $\ret{t}$ to be pseudo-concave and submodular. We now describe how costs are incorporated. This is done by using, instead of $\phr{\ret{t}}$, the function $\obf{t}$ defined by: 
\be
\obf{t}(\apv)=\npf{\ret{t}}(\apv)-\siz\sum_{i\in[\nmb]}\apc{i}\ctc{t}{i}
\ee
so that $\obf{t}(\pvt{t})$ lower bounds the expected profit on trial $t$. Since $\obf{t}$ differs from $\phr{\ret{t}}$ by a linear function it is straightforward to extend the above methodology to this new objective function.

%% file: sum-max.tex
\section{SUM-MAX FUNCTIONS}
\label{sec:sum-max}
We now introduce sum-max functions and define the key property of this class that allows us to learn it with bandit feedback.
\begin{definition}\label{smaxdef}
A set function $\af:2^{[\nmb]}\rightarrow \br$ is \emph{sum-max} if and only if there exists some $\nmu\in\nat$ and some matrix $\bs{V}\in\br^{\nmu\times\nmb}$ such that for all $\ass\subseteq[\nmb]$ with $\ass\neq\emptyset$ we have:
\be
\af(\ass)=\sum_{k\in[\nmu]}\max_{i\in\ass}V_{k,i}
\quad\text{and}\quad
\af(\emptyset)\leq\sum_{k\in[\nmu]}\min_{i\in[\nmb]}V_{k,i}
\ee
\end{definition}
For example, consider a marketplace with $\nmu$ buyers and $\nmb$ sellers. The value $V_{k,i}$ is the combined utility of buyer $k$ going to seller $i$. The value $\af(\ass)$ is the social welfare when a subset $\ass$ of sellers participate in the marketplace, and buyers match up with sellers to optimize their combined utilities. When there is only one buyer ($\nmu=1$), $\bs{V}$ is a vector $(V_1,\ldots,V_{\nmb})$ and we view each $i\in [\nmb]$ as an arm with reward $V_{i}$. Then $\af(\ass) = \max_{i\in\ass} V_i$, the maximum reward of an arm in the chosen set $\ass$.

As sum-max functions are sums of monotone submodular functions, they are monotone submodular. We now list a number of sequential decision-making problems that can be expressed as regret minimization of specific sum-max functions under bandit feedback.

\textbf{Bandit facility location.} This is the bandit version of the online facility location problem studied by \citet{Pasteris2020OnlineLO}. The (net) reward function then takes the form
\[
  \sum_{k\in[\nmu]}\max_{i\in\ass}V_{k,i} - \sum_{i\in\ass}\ctc{t}{i}
\]
We can view this setting as a generalization of the marketplace example where sellers pay a known cost to enter the market. At each round, the platform admits a subset $\ass$ of sellers and only observes the resulting social welfare (bandit feedback).

The following problems have zero costs. Hence, we impose a cardinality constraint $|\cho{t}| \le \siz$ on the learner's decision $\cho{t}$ at each time $t$.

\textbf{The multichannel campaign problem.} This is our nonstochastic variant of the best-of-$k$ bandit problem of \citet{simchowitz2016best}. To view it as an instance of sum-max optimization, set $\nmu=1$ and let $V_i\in\{0,1\}$ indicate whether a user makes a purchase when the ad is displayed on channel $i$. Then $(V_1,\ldots,V_{\nmb})$ can be viewed as the incidence vector of a subset $\aws{}\subseteq[\nmb]$ of channels, and the reward is defined by $\af(\ass)=\indi{\ass\cap\aws{}\neq\emptyset}$. The feedback is bandit because we do not know what channel triggered the sale for that user.

\textbf{Bandit hitting sets.} This is a generalization of the previous example where $\nmu \ge 1$ and $\bs{V}$ is a boolean matrix. Each row of $\bs{V}$ denotes a subset $\mathcal{C}_k$ of $[\nmb]$ and $V_{k,i}$ indicates whether $i \in \mathcal{C}_k$. The value $r(\ass)$ then counts how many sets $\mathcal{C}_k$ have a non-empty intersection with $\ass$. Bandit setting occurs when 
the sets remain unknown and each time we only observe the number of intersected sets.

\textbf{Combinatorial bandits.} Another important special case is when we receive the sum of the rewards $r_i$ of the arms $i\in\ass$. In this case $\nmu=\nmb$ and $V_{k,i}=\indi{k=i}r_i$. The problem is then equivalent to a combinatorial bandit (with full bandit feedback) over the class of $\siz$-sized subsets \citep{cesa2012combinatorial}.


\textbf{Bandit $k$-medians.} Given $\ave_1,\ldots,\ave_{\nmu}$ points in a metric space $(\mathcal{X},d)$, consider the version of the $k$-medians problem (for $k = \siz$) where the $\siz$ centroids have to be chosen in the given set of points. The value of the objective function at a candidate solution $\ass\subset [\nmb]$ with $|\ass|\leq\siz$ can be written as
\[
    \af(\ass) = -\sum_{k\in[\nmu]} \min_{i\in\ass} d(\ave_k,\ave_i)\,.
\]
Clearly, this is 
a sum-max function for $\bs{V}$ with elements $V_{k,i} := 
-d(\ave_k,\ave_i)
$. 
The feedback is bandit when we do not know the metric, but we can observe the value of the objective function.

Next, we introduce an important property of sum-max functions.
\begin{definition}\label{qcdef}
Suppose we have a set function $\af:2^{[\nmb]}\rightarrow\br$. For any $\ass\subseteq[\nmb]$ define the matrix $\mat{\af}{\ass}\in\br^{\nmb\times\nmb}$ such that
$
\mac{\af}{\ass}{i}{j}=\af(\ass\cup\{i,j\})
$
for all $i,j\in[\nmb]$.
We call the function $\af$ \emp{pseudo-concave} if and only if
$
\tra{\ave}\mat{\af}{\ass}\ave\leq0
$
for all $\ass\subseteq[\nmb]$ and all $\ave\in\br^{\nmb}$ with $\ave\cdot\bs{1}=0$.
\end{definition}
%

In Appendix \ref{apx:notpseudoconc}, we show that there are monotone submodular functions that are not pseudo-concave. As a consequence, sum-max functions are indeed a proper subset of the class of monotone submodular functions. The following theorem 
confirms that all sum-max functions are pseudo-concave:
\begin{restatable}{theorem}{summaxth}\label{summaxth}
Any sum-max set function is pseudo-concave.
\end{restatable}

\begin{proof} 
Suppose we have some sum-max function $\af:2^{[\nmb]}\rightarrow[0,1]$. Let $\bs{V}$ be as in Definition \ref{smaxdef}. Without loss of generality, we will assume that all components of $\bs{V}$ are non-negative and $\af(\emptyset)=0$ (since any sum-max function can be transformed into this form by the addition of a constant).\\
Define, for any $\bas\subseteq[\nmb]$, the set function $\af^{\bas}:2^{[\nmb]}\rightarrow[0,1]$ such that for all $\ass\subseteq[\nmb]$ we have
\be
\af^{\bas}(\ass):=\indi{\ass\cap\bas\neq\emptyset}
\ee
We shall now show that for all such $\bas$ we have that $\af^{\bas}$ is pseudo-concave. Choose any $\ave\in\br^{\nmb}$ with $\ave\cdot\bs{1}=0$ and any $\ass\subseteq[\nmb]$. We have two cases:
\begin{enumerate}[nosep,wide]
\item If $\ass\cap\bas\neq\emptyset$, for all $i,j\in[\nmb]$ we have $\af^{\bas}(\ass\cup\{i,j\})=1$, this implies
$\mat{\af^{\bas}}{\ass}=\bs{1}\tra{\bs{1}}$ and hence $
\tra{\ave}\mat{\af^{\bas}}{\ass}\ave=0
$
\item If $\ass\cap\bas=\emptyset$ then for all $i,j\in[\nmb]$ we have \be
\af^{\bas}(\ass\cup\{i,j\})=\indi{(i\in\bas)\vee(j\in\bas)}
\ee
Let $\bs{z}\in\{0,1\}^\nmb$ be such that for all $k\in[\nmb]$ we have $z_k:=\indi{k\notin\bas}$. Then for all $i,j\in[\nmb]$ we have
\be
\indi{(i\in\bas)\vee(j\in\bas)}=1-z_iz_j
\ee
so that, by above, we have
$
\mat{\af^{\bas}}{\ass}=\bs{1}\tra{\bs{1}}-\bs{z}\tra{\bs{z}}
$
This implies that:
$
\tra{\ave}\mat{\af^{\bas}}{\ass}\ave=-(\ave\cdot\bs{z})^2\leq0
$
\end{enumerate}
And therefore, $\af^\bas$ is pseudo-concave.

Now suppose we have a vector $\bs{v}\in\br_+^{\nmb}$ and define the set function 
$\af^{\bs{v}}:2^{[\nmb]}\rightarrow\br_+$ such that for all $\ass\subseteq[\nmb]$ we have
\be
\af^{\bs{v}}(\ass):=\max_{i\in\ass}v_i
\ee
where the maximum of the empty set is defined as equal to zero. We can order the set $[\nmb]$ into the sequence $\langle j_i\,|\,i\in[\nmb]\rangle=[\nmb]$ where $v_{j_{i+1}}\leq v_{j_i}$ for all $i\in[\nmb-1]$. For all $i\in[\nmb]$ we can define
$
\bas_i:=\{j_k\,|\,k\leq i\}
$. Now note then that for all $\ass\subseteq[\nmb]$ the set function $\af^{\bs{v}}(\ass)$ can be expressed as
\begin{align*}
    \sum_{i\in[\nmb-1]}&(v_{j_i} - v_{j_{i+1}})\indi{\ass\cap\bas_i\neq\emptyset}
    + v_{j_{\nmb}}\indi{\ass\cap\bas_\nmb\neq\emptyset}
\\&=
    \sum_{i\in[\nmb-1]}(v_{j_i}-v_{j_{i+1}})\af^{\bas_i}(\ass)+v_{j_{\nmb}}\af^{\bas_\nmb}(\ass)\,,
\end{align*}
so, by above, $\af^{\bs{v}}$ is a positive sum of pseudo-concave functions and is hence itself pseudo-concave. Note also that $\af^{\bs{v}}$ is clearly submodular. Noting that $\af$ is a positive sum of functions of the form $\af^{\bs{v}}$ we have now shown that it is both pseudo-concave and submodular as required.
\end{proof}

%% file: related.tex
\section{ADDITIONAL RELATED WORK}
The work closest to ours is \citet{Pasteris2020OnlineLO}, where they study online facility location with full information feedback. Our work improves on theirs in many respects: First, we solve the problem with bandit feedback, which requires designing an entirely different algorithm based on our discovery of an unbiased estimator for the gradient of our expected reward (we find it remarkable that such an estimator exists).
As a consequence, our algorithm is also applicable to the full-information setting, where 
we obtain a per-trial running time of $\mathcal{O}(\siz\nmb)$ when given an oracle for the reward function. When considering general sum-max functions, the methodology of \citet{Pasteris2020OnlineLO} would instead require a per-trial running time exponential in $\nmb$\footnote{The work of \citet{Pasteris2020OnlineLO} only considered single-user cases, but it is straightforward to extend their methodology to general sum-max functions.}. Second, our algorithm can efficiently learn classes that are even more general than sum-max functions. Third, we obtain tighter approximation ratios and show optimality for the multichannel campaign problem (and thus optimality in general).

Sum-max functions are a special case of linear submodular functions \citep{yue2011linear}, which are of the form $r(\ass) = \sum_{i \in [N]} w_i F_i(\ass)$ for $F_1,\ldots,F_N$ monotone submodular functions and $w_1,\ldots,w_N$ non-negative coefficients. However, linear submodular functions have been only studied in stochastic settings, assuming preliminary knowledge of $F_1,\ldots,F_N$, and using a feedback model more informative than our bandit feedback. 


Click-models \citep{lattimore2020bandit,lattimore2018toprank,kveton2015cascading} provide a different stochastic formalization of the best-of-$k$ bandit problem. Here the user is presented with an ordered list of items, and the learner receives a positive reward if the user clicks on one of the presented items. The difference with our multichannel campaign problem is that the items are ordered, and the likelihood of clicking an item is also affected by the position of the item within the list.

%% file: supplemental.tex
\appendix
\onecolumn

\section{ANALYSIS PROOFS (PROOF OF THEOREM \ref{mainth})}

\subsection{Lemma \ref{lemma:subsetcomp}} \label{apx:subsetcomp}

\lemmasubsetcomp*

\begin{proof}
\nc{\vse}[1]{\mathcal{V}_{#1}}
\nc{\vsd}[1]{\tilde{r}'_{#1}}
\nc{\upr}[1]{v(#1)} 

For all $k\in[K]\cup\{0\}$ define $\vse{k}:=\{\ass\in2^{[\nmb]}\,|\,k\leq|\ass| \}$. We take the inductive hypothesis such that for all $k\in[K]\cup\{0\}$  there exists a     unique function $\vsd{k}:\vse{k}\rightarrow\mathbb{R}$ such that for all $\bas\in\vse{k}$ we have
\be
\af(\bas)=\sum_{\ass\in\vse{k}}\indi{\bas\subseteq\ass}\vsd{k}(\ass)\,.
\ee
We will prove the inductive hypothesis via reverse induction on $k$ (i.e., from $k=K$ to $k=0$).

The inductive hypothesis holds for $k=K$ since the only element of $\vse{K}$ is $[\nmb]$ so we must have $\vsd{K}([\nmb]):=\af([\nmb])$

Now suppose, for some $i\in[\nmb]$\,, the inductive hypothesis holds when $k=i$. Now consider the case that $k=i+1$. Note that for all $\bas\in\vse{i}$ and $\ass\in\vse{i+1}\setminus\vse{i}$ we must have that $\bas\not\subseteq\ass$ and hence we must have that
\be
\af(\bas)=\sum_{\ass\in\vse{i}}\indi{\bas\subseteq\ass}\vsd{i+1}(\ass)\,,
\ee
so, by the inductive hypothesis, the restriction of $\vsd{i+1}$ to $\vse{i}$ is equal to $\vsd{i}$. Now choose some arbitrary $\bas\in\vse{i+1}\setminus\vse{i}$ and define:
\be
\upr{\bas}:=\sum_{\ass\in\vse{i}}\indi{\bas\subseteq{\ass}}\vsd{i+1}(\ass)
\ee
which, by above, is uniquely defined.
Note that for all $\ass\in\vse{i+1}\setminus\vse{i}$ we have that $\bas\subseteq\ass$ if and only if $\ass=\bas$ and hence we have that
\be
\af(\bas)=\sum_{\ass\in\vse{i}}\indi{\bas\subseteq{\ass}}\vsd{i+1}(\ass)+\vsd{i+1}(\bas)=\upr{\bas}+\vsd{i+1}(\bas)\,,
\ee
so that $\vsd{i+1}(\bas)=\af(\bas)-\upr{\bas}$ which is unique.

We have hence shown that the inductive hypothesis holds for $k=i+1$ and hence holds always. Noting that $\vse{0}=[\nmb]$ we then get the result by necessarily setting $\did=\vsd{0}$.
\end{proof}

\subsection{Lemma \ref{pheqexrelem}} \label{apx:pheqexrelem}

\pheqexrelem*

\begin{proof}
Let $\did$ be a subset decomposition of $\af$. We have
\begin{align*} 
\expt{\af(\asq{\apv})}&=\sum_{\ass\subseteq[\nmb]}\did(\ass)\prob{\asq{\apv}\subseteq\ass}\\
&=\sum_{\ass\subseteq[\nmb]}\did(\ass)\prod_{j\in[\siz]}\prob{\acq{j}{\apv}\in\ass}\\
&=\sum_{\ass\subseteq[\nmb]}\did(\ass)\prod_{j\in[\siz]}\sum_{i\in[\nmb]}\indi{i\in\ass}\apc{i}\\
&=\sum_{\ass\subseteq[\nmb]}\did(\ass)\left(\sum_{i\in[\nmb]}\indi{i\in\ass}\apc{i}\right)^\siz\\
&=\nps{\af}{\apv}
\end{align*}
as required.
\end{proof}

\subsection{Lemma \ref{derphilem1}}\label{apx:derphilem1}
\derphilem*

\begin{proof}
Let $\did$ be a subset decomposition of $\af$. For all $\apv'\in\br^{\nmb}$ and $\ass\subseteq[\nmb]$ define
\be
\nlm{\ass}{\apv'}:=\left(\sum_{k\in[\nmb]}\indi{k\in\ass}\apc{k}'\right)^{\siz}\,.
\ee
Fix some $j\in[\siz]$. Note that
\begin{align*}
\pder{i}\nlm{\ass}{\apv}&=\siz\indi{i\in\ass}\left(\sum_{k\in[\nmb]}\indi{k\in\ass}\apc{k}\right)^{\siz-1}\\
&=\siz\indi{i\in\ass}\prod_{j'\in[\siz]\setminus\{j\}}\sum_{k\in[\nmb]}\indi{k\in\ass}\apc{k}\\
&=\siz\indi{i\in\ass}\prod_{j'\in[\siz]\setminus\{j\}}\prob{\acq{j'}{\apv}\in\ass}\\
&=\frac{\siz}{\apc{i}}\prob{\acq{j}{\apv}=i}\indi{i\in\ass}\prod_{j'\in[\siz]\setminus\{j\}}\prob{\acq{j'}{\apv}\in\ass}\\
&=\frac{\siz}{\apc{i}}\prob{(\acq{j}{\apv}=i)\,\wedge\,(i\in\ass)}\prod_{j'\in[\siz]\setminus\{j\}}\prob{\acq{j'}{\apv}\in\ass}\\
&=\frac{\siz}{\apc{i}}\prob{(\acq{j}{\apv}=i)\,\wedge\,(\acq{j}{\apv}\in\ass)}\prod_{j'\in[\siz]\setminus\{j\}}\prob{\acq{j'}{\apv}\in\ass}\\
&=\frac{\siz}{\apc{i}}\prob{(\acq{j}{\apv}=i)\,\wedge\,(\forall j'\in[\siz]\,,\,\acq{j'}{\apv}\in\ass)}\\
&=\frac{\siz}{\apc{i}}\prob{(\acq{j}{\apv}=i)\,\wedge\,(\asq{\apv}\subseteq\ass)}\\
&=\frac{\siz}{\apc{i}}\expt{\indi{\acq{j}{\apv}=i}\indi{\asq{\apv}\subseteq\ass}}\,,
\end{align*}
so since:
\be
\nps{\af}{\apv}=\sum_{\ass\subseteq[\nmb]}\did(\ass)\nlm{\ass}{\apv}\,,
\ee
we have
\begin{align*}
\pder{i}\nps{\af}{\apv}&=\sum_{\ass\subseteq[\nmb]}\did(\ass)\pder{i}\nlm{\ass}{\apv}\\
&=\frac{\siz}{\apc{i}}\sum_{\ass\subseteq[\nmb]}\did(\ass)\expt{\indi{\acq{j}{\apv}=i}\indi{\asq{\apv}\subseteq\ass}}\\
&=\frac{\siz}{\apc{i}}\bexpt{\indi{\acq{j}{\apv}=i}\sum_{\ass\subseteq[\nmb]}\did(\ass)\indi{\asq{\apv}\subseteq\ass}}\\
&=\frac{\siz}{\apc{i}}\expt{\indi{\acq{j}{\apv}=i}\af(\asq{\apv})}\,.
\end{align*}
Summing over all $j\in[\siz]$ and dividing by $\siz$ then gives us
\be
\pder{i}\nps{\af}{\apv}=\bexpt{\frac{\af(\asq{\apv})}{\apc{i}}\sum_{j\in[\siz]}\indi{\acq{j}{\apv}=i}}
\ee
as required.
\end{proof}

\subsection{Lemma \ref{phiconlem1}} \label{apx:phiconlem1}
\phiconlem*

\begin{proof}
Choose any $\apv\in\splx{\nmb}$. Define $\langle\acqp{j}{\apv}\,|\,j\in[\siz-2]\rangle$ to be a sequence of stochastic quantities drawn i.i.d. at random from (the probability distribution characterised by) $\apv$. In addition, let:
\be
\asqp{\apv}:=\{\acqp{j}{\apv}\,|\,j\in[\siz-2]\}
\ee
Direct from the definition of $\npf{\af}$ we have, for all $i,i'\in[\nmb]$, that
\begin{align*}
\pder{i}\pder{i'}\npf{\af}&=\sum_{\ass\subseteq[\nmb]}\did(\ass)\indi{i\in\ass}\indi{i'\in\ass}\left(\sum_{k\in[\nmb]}\indi{k\in\ass}\apc{k}\right)^{\siz-2}\\
&=\sum_{\ass\subseteq[\nmb]}\did(\ass)\indi{i\in\ass}\indi{i'\in\ass}\prod_{j\in[\siz-2]}\sum_{k\in[\nmb]}\indi{k\in\ass}\apc{k}\\
&=\sum_{\ass\subseteq[\nmb]}\did(\ass)\indi{i\in\ass}\indi{i'\in\ass}\prod_{j\in[\siz-2]}\prob{\acqp{j}{\apv}\in\ass}\\
&=\sum_{\ass\subseteq[\nmb]}\did(\ass)\indi{i\in\ass}\indi{i'\in\ass}\prob{\asqp{\apv}\subseteq\ass}\\
&=\sum_{\ass\subseteq[\nmb]}\did(\ass)\prob{\asqp{\apv}\cup\{i,i'\}\subseteq\ass}\\
&=\bexpt{\sum_{\ass\subseteq[\nmb]}\did(\ass)\indi{\asqp{\apv}\cup\{i,i'\}\subseteq\ass}}\\
&=\expt{\af(\asqp{\apv}\cup\{i,i'\})}\\
&=\sum_{\ass\subseteq[\nmb]}\prob{\asqp{\apv}=\ass}\af(\ass\cup\{i,i'\})\\
&=\sum_{\ass\subseteq[\nmb]}\prob{\asqp{\apv}=\ass}\mac{\af}{\ass}{i}{i'}\,.
\end{align*}
So for all $\ave\in\br^{\nmb}$ with $\ave\cdot\bs{1}=0$ we have
\begin{align*}
\tra{\ave}(\nabla^2\npf{\af}(\apv))\ave&=\sum_{i,i'\in[\nmb]}\avec{i}(\pder{i}\pder{i'}\npf{\af})\avec{i'}\\
&=\sum_{i,i'\in[\nmb]}\avec{i}\avec{i'}\sum_{\ass\subseteq[\nmb]}\prob{\asqp{\apv}=\ass}\mac{\af}{\ass}{i}{i'}\\
&=\sum_{\ass\subseteq[\nmb]}\prob{\asqp{\apv}=\ass}\sum_{i,i'\in[\nmb]}\avec{i}\mac{\af}{\ass}{i}{i'}\avec{i}\\
&=\sum_{\ass\subseteq[\nmb]}\prob{\asqp{\apv}=\ass}(\tra{\ave}\mat{\af}{\ass}\ave)\\
&\leq0\,,
\end{align*}
which means that $\npf{\af}$ is concave on $\splx{\nmb}$ as required.
\end{proof}

\subsection{Lemma \ref{psubrlem1}}\label{apx:psubrlem2}

\psubrlem*

\begin{proof}
Without loss of generality assume that $\af(\emptyset)=0$. 

We prove by induction on $\ltm$ that the lemma holds whenever $\siz\leq\ltm$. In the case that $\ltm=0$ we have $\expt{\af(\asr)}=\af(\emptyset)=0$ and $\siz=0$ so the result holds. Now assume that it holds for all $\siz\leq\ltm$ and consider the case that $\siz=\ltm+1$.

Since $\af$ is pseudo-submodular choose $i\in\ass$ such that
\begin{equation}\label{choieq1}
\af(\bas\cup\{i\})-\af(\bas)\geq\af(\ass)-\af(\ass\setminus\{i\})
\end{equation}
for all $\bas\subseteq\ass\setminus\{i\}$. Define $\cas:=|\ass|$ and 
\begin{equation}\label{choieq2}
\dif:=\af(\ass)-\af(\ass\setminus\{i\})\,.
\end{equation}
Let $\langle\lz{s}\,|\,s\in[\siz]\rangle$ be a sequence of $\siz$ elements drawn uniformly at random from $\ass$ such that $\asr=\{\lz{s}\,|\,s\in[\siz]\}$. Define
\be
\nni:=\sum_{s\in[\siz]}\indi{\lz{s}\neq i}\,.
\ee
For all $j\in[\siz]\cup\{0\}$ let $\asm{j}$ be a set formed by sampling $j$ actions 
independently and uniformly at random from $\ass\setminus\{i\}$.

Note that by the inductive hypothesis, we have
\begin{align}
\notag\expt{\indi{i\notin\asr}\af(\asr)}&=\prob{i\notin{\asr}}\cex{\af(\asr)}{i\notin\asr}\\
\notag&=\prob{i\notin{\asr}}\left(1-\left(\frac{|\ass\setminus\{i\}|-1}{|\ass\setminus\{i\}|}\right)^\siz\right)\af(\ass\setminus\{i\})\\
\notag&=\prob{i\notin{\asr}}\left(1-\left(\frac{\cas-2}{\cas-1}\right)^\siz\right)\af(\ass\setminus\{i\})\\
\label{choieq7}&=\prob{\nni=\siz}\left(1-\left(\frac{\cas-2}{\cas-1}\right)^\siz\right)\af(\ass\setminus\{i\})
\end{align}
Note also that
\begin{equation}\label{choieq5}
\expt{\indi{i\in\asr}\af(\asr)}=\sum_{j\in[\ltm]\cup\{0\}}\prob{\nni=j}\expt{\af(\asm{j}\cup\{i\})}\,.
\end{equation}
By equations \eqref{choieq1} and \eqref{choieq2} and the inductive hypothesis we have, for all $j\in[\ltm]\cup\{0\}$, that
\begin{align}
\notag\expt{\af(\asm{j}\cup\{i\})}&\geq\expt{\dif+\af(\asm{j})}\\
\notag&=\dif+\expt{\af(\asm{j})}\\
\notag&\geq\dif+\left(1-\left(\frac{|\ass\setminus\{i\}|-1}{|\ass\setminus\{i\}|}\right)^j\right)\af(\ass\setminus\{i\})\\
\label{choieq3}&=\dif+\left(1-\left(\frac{\cas-2}{\cas-1}\right)^j\right)\af(\ass\setminus\{i\})\,.
\end{align}
We also have that
\begin{equation}\label{choieq4}
\sum_{j\in[\ltm]\cup\{0\}}\prob{\nni=j}=\prob{i\in\asr}\,.
\end{equation}
Substituting equations \eqref{choieq3} and \eqref{choieq4} into Equation \eqref{choieq5} gives us
\be
\expt{\indi{i\in\asr}\af(\asr)}=\prob{i\in\asr}\dif+\sum_{j\in[\ltm]\cup\{0\}}\prob{\nni=j}\left(1-\left(\frac{\cas-2}{\cas-1}\right)^j\right)\af(\ass\setminus\{i\})\,.
\ee
Adding this equation to Equation \eqref{choieq7} gives us
\begin{equation}\label{choieq8}
\expt{\af(\asr)}=\prob{i\in\asr}\dif+\sum_{j\in[\siz]\cup\{0\}}\prob{\nni=j}\left(1-\left(\frac{\cas-2}{\cas-1}\right)^j\right)\af(\ass\setminus\{i\})\,.
\end{equation}
Take any $k\in\ass\setminus\{i\}$. Note that
\be
1-\left(\frac{\cas-2}{\cas-1}\right)^j=1-(1-1/(\cas-1))^j=1-\prob{k\notin\asm{j}}=\prob{k\in\asm{j}}\,,
\ee
so that
\begin{align*}
\sum_{j\in[\siz]\cup\{0\}}\prob{\nni=j}\left(1-\left(\frac{\cas-2}{\cas-1}\right)^j\right)&=\sum_{j\in[\siz]\cup\{0\}}\prob{\nni=j}\prob{k\in\asm{j}}\\
&=\sum_{j\in[\siz]\cup\{0\}}\prob{\nni=j}\prob{k\in\asr\setminus\{i\}\,|\,\nni=j}\\
&=\sum_{j\in[\siz]\cup\{0\}}\prob{\nni=j}\prob{k\in\asr\,|\,\nni=j}\\
&=\prob{k\in\asr}\,.
\end{align*}

Substituting into Equation \eqref{choieq8} gives us:
\begin{align*}
\expt{\af(\asr)}&\geq\prob{i\in\asr}\dif+\prob{k\in\asr}\af(\ass\setminus\{i\})\\
&=\prob{i\in\asr}(\dif+\af(\ass\setminus\{i\}))\\
&=\prob{i\in\asr}\af(\ass)\\
&=(1-\prob{i\notin\asr})\af(\ass)\\
&=(1-(1-1/\cas)^\siz)\af(\ass)\\
&=\left(1-\left(\frac{|\ass|-1}{|\ass|}\right)^\siz\right)\af(\ass)\,.
\end{align*}
So the inductive hypothesis holds for all $\siz\in[\ltm+1]$ and hence holds always.
\end{proof}

\subsection{Lemma \ref{expslem}} \label{apx:expslem}

\expslem*

\begin{proof}
Let $\asr$ be a set formed by drawing $\siz$ elements i.i.d. with replacement from $\ass$. Let $\asre$ be an element drawn i.i.d. from $\ass$. Let $\did$ be a subset-decomposition of $\af$. Note that for all $i\in[\nmb]$ we have
\be
\psti{\ass}{i}=\prob{\asre=i}\,.
\ee
Hence, we have
\begin{align*}
\npf{\af}(\pstv{\ass})&=\sum_{\bas\subseteq[\nmb]}\did(\bas)\left(\sum_{i\in[\nmb]}\indi{i\in\bas}\psti{\ass}{i}\right)^\siz\\
&=\sum_{\bas\subseteq[\nmb]}\did(\bas)\left(\sum_{i\in[\nmb]}\indi{i\in\bas}\prob{\asre=i}\right)^\siz\\
&=\sum_{\bas\subseteq[\nmb]}\did(\bas)\prob{\asre\in\bas}^\siz\\
&=\sum_{\bas\subseteq[\nmb]}\did(\bas)\prob{\asr\subseteq\bas}\\
&=\sum_{\bas\subseteq[\nmb]}\did(\bas)\expt{\indi{\asr\subseteq\bas}}\\
&=\bexpt{\sum_{\bas\subseteq[\nmb]}\did(\bas)\indi{\asr\subseteq\bas}}\\
&=\expt{\af(\asr)}
\end{align*}
So
\be
\npf{\af}(\pstv{\ass})-\af(\emptyset)=\expt{\af(\asr)-\af(\emptyset)}\,,
\ee
Lemma \ref{psubrlem1} then gives us the result.
\end{proof}

\subsection{Lemma \ref{unbgreslem1}} \label{apx:unbgreslem1}

\unbgreslem*

\begin{proof}
Take any $i\in[\nmb]$. For any $j\in[\siz]$ we have
\begin{align*}
\ctc{t}{i}&=\pct{t}{i}\ctc{t}{i}/\pct{t}{i}\\
&=\prob{\acc{t}{j}=i\,|\,\pvt{t}}\ctc{t}{i}/\pct{t}{i}\\
&=\cex{\indi{\acc{t}{j}=i}\ctc{t}{i}/\pct{t}{i}}{\pvt{t}}
\end{align*}
so:
\begin{align*}
\siz\ctc{t}{i}&=\sum_{j\in[\siz]}\cex{\indi{\acc{t}{j}=i}\ctc{t}{i}/\pct{t}{i}}{\pvt{t}}\\
&=\cbex{\frac{\ctc{t}{j}}{\pct{t}{i}}\sum_{j\in[\siz]}\indi{\acc{t}{j}=i}}{\pvt{t}}
\end{align*}
Hence, by Lemma \ref{derphilem1}, we have
\begin{align*}
\pder{i}\obf{t}(\pvt{t})&=\pder{i}\npf{\ret{t}}(\pvt{t})-\siz\ctc{t}{i}\\
&=\bexpt{\frac{\ret{t}(\asq{\pvt{t}})}{\pct{t}{i}}\sum_{j\in[\siz]}\indi{\acq{j}{\pvt{t}}=i}}-\siz\ctc{t}{i}\\
&=\cbex{\frac{\ret{t}(\cho{t})}{\pct{t}{i}}\sum_{j\in[\siz]}\indi{\acc{t}{j}=i}}{\pvt{t}}-\siz\ctc{t}{i}\\
&=\cex{\egr{t}{i}}{\pvt{t}}
\end{align*}
as required.
\end{proof}

\subsection{Lemma \ref{susqlem}}\label{apx:susqlem}

\susqlem*

\begin{proof}
Given $i\in[\nid]$ we have that
\begin{align*}
\frac{\expt{\egr{t}{i}^2}}{(1+\mxc)^2}&=\frac{1}{(1+\mxc)^2}\bexpt{(\ret{t}(\cho{t})-\ctc{t}{i})^2\sum_{j,j'\in[\siz]}\frac{\indi{\acc{t}{j}=i}\indi{\acc{t}{j'}=i}}{\pct{t}{i}^2}}\\
&\leq\sum_{j,j'\in[\siz]}\bexpt{\frac{\indi{\acc{t}{j}=i}\indi{\acc{t}{j'}=i}}{\pct{t}{i}^2}}\\
&=\sum_{j\in[\siz]}\bexpt{\frac{\indi{\acc{t}{j}=i}}{\pct{t}{i}^2}}+\sum_{j,j'\in[\siz]}\indi{j\neq j'}\bexpt{\frac{\indi{\acc{t}{j}=i}\indi{\acc{t}{j'}=i}}{\pct{t}{i}^2}}\\
&=\sum_{j\in[\siz]}\frac{\prob{\acc{t}{j}=i}}{\pct{t}{i}^2}+\sum_{j,j'\in[\siz]}\indi{j\neq j'}\frac{\prob{\acc{t}{j}=i}\prob{\acc{t}{j'}=i}}{\pct{t}{i}^2}\\
&=\sum_{j\in[\siz]}\frac{1}{\pct{t}{i}}+\sum_{j,j'\in[\siz]}\indi{j\neq j'}\\
&=\frac{\siz}{\pct{t}{i}}+\siz(\siz-1)\,,
\end{align*}
and hence
\be
\bexpt{\sum_{i\in[\nmb]}\pct{t}{i}\egr{t}{i}^2}=\sum_{i\in[\nmb]}\pct{t}{i}\expt{\egr{t}{i}^2}\leq(1+\mxc)^2\siz(\nmb+\siz-1)
\ee
as required.
\end{proof}

\subsection{Theorem \ref{mainth}} \label{apx:final_lemma}
\mainth*

\begin{proof}
Consider some trial $t\in[\ntr]$. By Lemma \ref{phiconlem1} and the definition of $\obf{t}$ we have that $\obf{t}$ is concave over $\splx{\nmb}$. Hence, by Lemma \ref{unbgreslem1}, we have
\begin{align}
\notag\cex{(\pstv{\ass}-\pvt{t})\cdot\gra{t}}{\pvt{t}}&=(\pstv{\ass}-\pvt{t})\cdot\cex{\gra{t}}{\pvt{t}}\\
\notag&=(\pstv{\ass}-\pvt{t})\cdot\der\obf{t}(\pvt{t})\\
\label{filemdogreq1}&\geq\obf{t}(\pstv{\ass})-\obf{t}(\pvt{t})
\end{align}
Lemma \ref{expslem} gives us:
\begin{align}
\notag\obf{t}(\pstv{\ass})&=\npf{\ret{t}}(\pstv{\ass})-\siz\sum_{i\in[\nmb]}\psti{\ass}{i}\ctc{t}{i}\\
\label{filemdogreq2}&\geq\af(\emptyset)+\left(1-\left(\frac{|\ass|-1}{|\ass|}\right)^\siz\right)(\af(\ass)-\af(\emptyset))-\frac{\siz}{|\ass|}\sum_{i\in\ass}\ctc{t}{i}
\end{align}
and Lemma \ref{pheqexrelem} gives us:
\begin{align}
\notag\obf{t}(\pvt{t})&=\npf{\ret{t}}(\pvt{t})-\siz\sum_{i\in[\nmb]}\pct{t}{i}\ctc{t}{i}\\
\notag&=\expt{\ret{t}(\asq{\pvt{t}})}-\siz\sum_{i\in[\nmb]}\pct{t}{i}\ctc{t}{i}\\
\notag&=\cex{\ret{t}(\cho{t})}{\pvt{t}}-\siz\sum_{i\in[\nmb]}\pct{t}{i}\ctc{t}{i}\\
\notag&=\cex{\ret{t}(\cho{t})}{\pvt{t}}-\sum_{j\in[\siz]}\sum_{i\in[\nmb]}\prob{\acc{t}{j}=i\,|\,\pvt{t}}\ctc{t}{i}\\
\notag&=\cex{\ret{t}(\cho{t})}{\pvt{t}}-\sum_{j\in[\siz]}\cex{\ctc{t}{\acc{t}{j}}}{\pvt{t}}\\
\notag&=\cex{\ret{t}(\cho{t})}{\pvt{t}}-\cbex{\sum_{j\in[\siz]}\ctc{t}{\acc{t}{j}}}{\pvt{t}}\\
\notag&\leq\cex{\ret{t}(\cho{t})}{\pvt{t}}-\cbex{\sum_{i\in\cho{t}}\ctc{t}{\acc{t}{j}}}{\pvt{t}}\\
\label{filemdogreq3}&=\cex{\pro{t}}{\pvt{t}}+\ret{t}(\emptyset)
\end{align}
Substituting equations \eqref{filemdogreq2} and \eqref{filemdogreq3} into Equation \eqref{filemdogreq1} gives us:
\be
\cex{(\pstv{\ass}-\pvt{t})\cdot\gra{t}}{\pvt{t}}\geq-\cex{\pro{t}}{\pvt{t}}+\left(1-\left(\frac{|\ass|-1}{|\ass|}\right)^\siz\right)\dir{t}(\ass)-\frac{\siz}{|\ass|}\sum_{i\in\ass}\ctc{t}{i}
\ee
and hence:
\begin{equation}\label{filemdogreq4}
\expt{(\pstv{\ass}-\pvt{t})\cdot\gra{t}}\geq-\expt{\pro{t}}+\left(1-\left(\frac{|\ass|-1}{|\ass|}\right)^\siz\right)\dir{t}(\ass)-\frac{\siz}{|\ass|}\sum_{i\in\ass}\ctc{t}{i}
\end{equation}
Lemma \ref{susqlem} gives us:
\begin{equation}\label{filemdogreq5}
\bexpt{\sum_{i\in[\nmb]}\pct{t}{i}\egr{t}{i}^2}\leq\frac{\rgt^2}{\ntr}
\end{equation}
Lemma \ref{hedgelem1} gives us:
\begin{equation}\label{filemdogreq6}
\sum_{t\in[\ntr]}\expt{(\pstv{\ass}-\pvt{t})\cdot\gra{t}}\leq\frac{\ln(\nmb)}{\lr}+\lr\sum_{t\in[\ntr]}\bexpt{\sum_{i\in[\nid]}\pct{t}{i}\egr{t}{i}^2}
\end{equation}
Substituting equations \eqref{filemdogreq4} and \eqref{filemdogreq5} into Equation \eqref{filemdogreq6} gives us:
\be
-\sum_{t\in[\ntr]}\expt{\pro{t}}+\left(1-\left(\frac{|\ass|-1}{|\ass|}\right)^\siz\right)\sum_{t\in[\ntr]}\dir{t}(\ass)-\frac{\siz}{|\ass|}\sum_{t\in[\ntr]}\sum_{i\in\ass}\ctc{t}{i}\leq\frac{\ln(\nmb)}{\lr}+\lr\rgt^2
\ee
Since $\lr=\ln(\nmb)/\rgt$ this implies the result.
\end{proof}

\section{PROOF OF THEOREM \ref{Th:LB1}}\label{apx:LB1}
\thLB*
\begin{proof}
Suppose we have such an algorithm. Let $\rct>0$ and $\cco>1$ be such that
\begin{equation}\label{arbeq2}
\bexpt{\sum_{t\in[\ntr]}\ret{t}(\cho{t})}\geq\left(1-\left(\frac{|\ars|-1}{|\ars|}\right)^{\cco\siz}\right)\sum_{t\in[\ntr]}\ret{t}(\ars)- \rct\nid^\kex\ntr^\tpo\,.
\end{equation}
Choose any $\rho\in(1/\cco,1)$ and then consider any $\ssi\in\nat$ such that $\ssi>4^{1/(\rho\cco-1)}$. Consider also any collection of sets $\{\cov{k}~|~k\in[\nid]\}\subseteq2^{[\ssi]}$ such that
\be
\bigcup_{k\in[\nid]}\cov{k}=[\ssi]\,.
\ee
Let $\mco$ be a subset of $[\nid]$ of minimum cardinality such that
\be
\bigcup_{k\in\mco}\cov{k}=[\ssi]\,.
\ee
Now choose
\be
\ntr:=\left\lceil(4\rct\nid^{\kex}\ssi)^{1/(1-\tpo)}\right\rceil\,.
\ee 
and choose any $\siz\in\nat$ such that $\siz\geq\rho\ln(\ssi)|\ars|$. For all $t\in[\ntr]$ draw $\aws{t}$ randomly as follows. First draw $\bet{t}$ uniformly at random from $[\ssi]$ and then define
\be
\aws{t}:=\{k\in[\nid]~|~\bet{t}\in\cov{k}\}\,.
\ee
It is a classic result that
\be
\left(\frac{|\ars|-1}{|\ars|}\right)^{|\ars|}\leq e^{-1}\,.
\ee
so by the conditions on $\ssi$ and $\siz$ we have
\begin{equation}\label{arbeq3}
\left(\frac{|\ars|-1}{|\ars|}\right)^{\cco\siz}\leq\exp(-\cco\siz/|\ars|)=\ssi^{-\rho\cco}=\frac{\ssi^{1-\rho\cco}}{\ssi}<\frac{1}{4\ssi}\,.
\end{equation}
By definition of $\mco$ we have, for all $t\in[\ntr]$, that there exists some $k\in\mco$ such that $\bet{t}\in\cov{k}$ so that $\aws{t}\cap\mco\neq\emptyset$. This implies
\be
\sum_{t\in[\ntr]}\ret{t}(\ars)=\ntr\,,
\ee
and hence, by~\eqref{arbeq2} and~\eqref{arbeq3}, we have
\begin{equation}\label{arbeq1}
\bexpt{\sum_{t\in[\ntr]}(1-\ret{t}(\cho{t}))}\leq\ntr-\ntr+\frac{\ntr}{4\ssi}+\rct\nid^\kex\ntr^\tpo\leq \frac{\ntr}{4\ssi} +\frac{\ntr\rct\nid^\kex}{\ntr^{1-\tpo}}\leq\frac{\ntr}{2\ssi}\,.
\end{equation}
Fix $t$ and a realization of $\cho{t}$. If we have
\be
\bigcup_{k\in\cho{t}}\cov{k}\neq[\ssi]\,,
\ee
then we must also have that
\begin{align*}
    \expt{1-\ret{t}(\cho{t}) \mid \cho{t}}
&=
    \prob{\cho{t}\cap\aws{t}=\emptyset \mid \cho{t}}
\\ &=
    \prob{\forall\, k\in\cho{t}\,,\,\bet{t}\notin\cov{k} \mid \cho{t}}
\\ &=
    \bprob{\left.\bet{t}\notin\bigcup_{k\in\cho{t}}\cov{k} \,\right\vert\, \cho{t}}
\ge
    \frac{1}{\ssi}\,.    
\end{align*}
Hence, by taking the randomness of $\cho{1},\ldots,\cho{T}$ into account,
\begin{align*}
    &\bprob{\sum_{t\in[\ntr]} \indi{\bigcup_{k\in\cho{t}}\cov{k}\neq[\ssi]} = T}
\\ &\le
    \bprob{\bexpt{\left.\sum_{t\in[\ntr]} \big(1-\ret{t}(\cho{t})\big) \,\right\vert\, \cho{1},\ldots,\cho{T}} \ge \frac{T}{n}}
\le
    \frac{1}{2}
\end{align*}
by \eqref{arbeq1}.
Since $\ntr$ is polynomial in $\nid\ssi$ and $|\cho{t}|\leq\siz$, we have a randomized polynomial-time algorithm that, with probability at least $\frac{1}{2}$, solves the set cover problem on $[\ssi]$ with approximation ratio $(1-\sca)\ln(\ssi)$ for $\sca = 1 - \rho \in \big(0,1-1/\cco\big)$.
\end{proof}

\section{SUBMODULAR MONOTONE NON-PSEUDOCONCAVE FUNCTIONS} \label{apx:notpseudoconc}

We provide a function counterexample to show that there are submodular monotone functions which are not pseudoconcave.

Let $K=8$, $\mathcal{P}=2^{[K]}$, $\mathcal{S}=\{K\}$, and $\alpha>0$. We define $U^{r,\mathcal{S}}$ as follows:

\[
U^{r,\mathcal{S}}:=\begin{pmatrix}
1 & 2 & 2 & 2 & 1+\alpha & 1+\alpha & 1+\alpha & 1 \\
2 & 1 & 2 & 2 & 1+\alpha & 1+\alpha & 1+\alpha & 1 \\
2 & 2 & 1 & 2 & 1+\alpha & 1+\alpha & 1+\alpha & 1 \\
2 & 2 & 2 & 1 & 1+\alpha & 1+\alpha & 1+\alpha & 1 \\
1+\alpha & 1+\alpha & 1+\alpha & 1+\alpha & 1 & 2 & 2 & 1 \\
1+\alpha & 1+\alpha & 1+\alpha & 1+\alpha & 2 & 1 & 2 & 1 \\
1+\alpha & 1+\alpha & 1+\alpha & 1+\alpha & 2 & 2 & 1 & 1 \\
1 & 1 & 1 & 1 & 1 & 1 & 1 & 0
\end{pmatrix}\,.
\]

\bigskip 

Now, let $\mathbf{x}=(1,1,1,1,-1,-1,-1,-1)^{\top}$. Note that we have $\langle\mathbf{x},\mathbf{1}\rangle=0$ as required by the pseudoconcavity definition, and $\mathbf{x}^{\top} U^{r,\mathcal{S}} \mathbf{x}=17-24\alpha$, which is positive for $\alpha\in\left(0,\frac{17}{24}\right)$, implying therefore the non-pseudoconcavity of $r$ for such values of $\alpha$.

\bigskip

We now show how to define $r$ starting from $U^{r,\mathcal{S}}$ in such a way that it is both monotone and submodular while being therefore also non-pseudoconcave. 

\smallskip

We have $|\mathcal{P}|=2^K=256$ possible subsets  as the arguments of $r$, $29$ of which are already defined by the above matrix $U^{r,\mathcal{S}}$: 
\begin{itemize}
    \item $1$ subset ($\{K\}$) with cardinality $1$,
    \item $7$ subsets ($\{j,K\}_{j\in [K-1]}$)  with cardinality $2$,
    \item $21$ subsets ($\{i,j,K\}_{1\le j<i\le K-1}$) with cardinality $3$.
\end{itemize}

For any $i\in [K]$, let $\delta_i$ and $\Delta_i$ be equal respectively to the minimum and the maximum difference (gain) over all values of $r$ for subsets with cardinality $i$ and all the ones for subsets with cardinality $i-1$. As previously anticipated, we construct function $r$ starting from the above matrix $U^{r,\mathcal{S}}$ in such a way that for all $i\in [K-1]$, we have
\[
\delta_{i}\ge\Delta_{i+1}\,,
\] 
which is a sufficient condition for submodularity because, for all $i\in [K]$, {\em each} subset $S_i\in\mathcal{P}$ with cardinality $i$ can be generated by adding one of its element {\em only} from a subset $S_{i-1}\subset S_i$ with cardinality $i-1$.

We set $\alpha=\frac{2}{3}<\frac{17}{24}$, which guarantees the non-pseudoconcavity of $r$.  To ensure monotonicity and submodularity, we define 
\begin{itemize}
    \item $r(S_1):=0$ for all subsets $S_1\in\mathcal{P}$ with $|S_1|=1$ (consistently with $U^{r,\mathcal{S}}_{K,K})$;
    \item $r(S_2):=1$ for all subsets $S_2\in\mathcal{P}$ with $|S_2|=2$ (consistently with $U^{r,\mathcal{S}}_{K,j}, U^{r,\mathcal{S}}_{j,j}, U^{r,\mathcal{S}}_{j,1}$ for all $j\in [K-1]$); 
    \item $r(S_3):=1+\frac{2}{3}=\frac{5}{3}$ for all subsets $S_3\in\mathcal{P}$ with $|S_3|=3$ that are not already defined by $U^{r,\mathcal{S}}$;
    \item $r(S_4):=r(S_3)+\frac{1}{2}=\frac{5}{3}+\frac{1}{2}=2+\frac{1}{6}> \max_{i,j} U^{r,\mathcal{S}}_{i,j}=2$ for all subsets $S_4\in\mathcal{P}$ with $|S_4|=4$;
    \item $r(S_5):=r(S_4)+\frac{1}{6}=2+\frac{2}{6}$,\\$r(S_6):=r(S_5)+\frac{1}{6}=2+\frac{3}{6}$,\\$r(S_7):=r(S_6)+\frac{1}{6}=2+\frac{4}{6}$, \\$r(S_8):=r(S_7)+\frac{1}{6}=2+\frac{5}{6}$\\for all subsets $S_5,S_6,S_7,S_8\in\mathcal{P}$ such that $|S_5|=5$, $|S_6|=6$, $|S_7|=7$, $|S_8|=8$.
\end{itemize}
Finally, we also set $r(\emptyset)=-1$. Note that, to ensure that submodularity is not violated, for each subset $S_3^U$ with $|S_3^U|=3$ defined by $U^{r,\mathcal{S}}$, we have that the difference $r(S_3^U)-r(S_2)$ for any subset $S_2\subset S_3^U$ with $|S_2|=2$ is either equal to $\alpha=\frac{2}{3}$ or $1$, that is not smaller than the maximum difference $r(S_4)-r(S_3)$ over all $S_3,S_4\in\mathcal{P}$ with $|S_3|=3$ and $|S_4|=4$, which in turn is equal to $\tfrac12<\tfrac23$. Furthermore, $r(S_4)=2+\frac{1}{6}$ is never smaller than any values of $r(S_3^U)$ for all subsets $S_3^U\in\mathcal{P}$ with $|S_3|=3$ that are already defined by $U^{r,\mathcal{S}}$, because we have $r(S_3^U)\le 2$, thereby preserving monotonicity for all subsets in $\mathcal{P}$ with cardinality smaller or equal to $4$.


Now, we recall that for any $i\in [K]$, $\delta_i$ and $\Delta_i$ are defined to be respectively equal to the minimum and the maximum difference (gain) over all values of $r$ for subsets with cardinality $i$ and all the ones for subsets with cardinality $i-1$. Since we have 
\begin{itemize}
    \item $\delta_1,\Delta_1,\delta_2,\Delta_2=1$ ~~(which immediately implies $\Delta_2\le\delta_1$),
    \item $\delta_3=\frac{2}{3};~~~\Delta_3=1\le\delta_2$,
    \item $\delta_4=\frac{1}{6};~~~\Delta_4=\frac{1}{2}\le\delta_3$,
    \item $\delta_5,\Delta_5,\delta_6,\Delta_6,\delta_7,\Delta_7,\delta_8,\Delta_8=\frac{1}{6}\le\delta_4$,
\end{itemize}

then  $\delta_{i}\ge\Delta_{i+1}$
 for all $i\in [K-1]$ which guarantees the submodularity of $r$. Finally, it is immediate to verify that $r$ is monotone also for all subsets in $\mathcal{P}$ with cardinality larger than $4$. Hence, we conclude that $r$ is monotone submodular and non-pseudoconcave. 

\hfill $\qedsymbol{}$

\section{EFFICIENT IMPLEMENTATION OF \textsc{COMBAND}} \label{apx:comband}

To implement the algorithm the \textsc{ComBand} presented in  \textit{\cite{cesa2012combinatorial}}, 
it is necessary to devise an efficient method for sampling from a set 
whose size can be exponential in $K$. In fact, at each trial, 
given a set $\mathcal{S}$ of positive real numbers, 
we need to select any of the subsets $S$ with a given size $m$ from $\mathcal{S}$ with a probability proportional to the product of the elements contained in $S$ itself.

To be consistent with the notation used in~\cite{cesa2012combinatorial}, henceforth we use the symbol $d$ in place of $K$.

Given a set $\mathcal{S}=\{q_1,q_2,\ldots,q_d\}$ 
of real positive numbers, we now show how to select a $m$-sized subset of indices $\{u_1,\ldots,u_m\}$ from $[d]$  with a probability proportional to $\Pi_{i=1}^m q_{u_i}$ by using dynamic programming. The running time of this sampling method is always linear\footnote{We assume that multiplying two numbers requires a constant time. Removing this assumption, since it is known that it is possible to multiply two numbers represented by at most $m$ bits in time equal to $\widetilde{\mathcal{O}}(m)$ when $m\gg 1$~\cite{harvey2021integer}, the total sampling time would be $\widetilde{\mathcal{O}}(m^2d)$  instead of $\mathcal{O}(md)$.} in $m\cdot d$.

For each sampling operation, consider the sequence of element indices $u_1, u_2, \ldots, u_m$ ordered according to the elements in $[d]$, i.e., $u_i<u_{i+1}$ for all $i\in [m-1]$. 

The main idea of this method is to sample first $u_m$, and then $u_{m-1},\ldots, u_1$ (i.e., in reverse order) having derived in a preliminary phase via dynamic programming all the probabilities that $u_m=j$ for all $m\le j\le d$, and the conditional probabilities that $u_{m'}=j$ given that $u_{m'+1}=j'$, for all $m'\in [m-1]$ and $m'\le j< j'\le d-m+m'$.

We denote the conditional probability that $u_{m'}=j$ given that $u_{m'+1}=j'$, where $m'\in [m-1]$ and $m'\le j< j'\le d-m+m'$ by 

\[
P_{m',j|j'}:=\mathbb{P}(u_{m'}=j|u_{m'+1}=j')\,,
\]

and, for the selection of $u_m$, we define for all $j\in [d]$

\[
P_{m,j}:=\mathbb{P}(u_{m}=j)\,,
\]

because there is no element $u_{j'}>u_m$ (with $j'>m$) in the sequence of selected indices from $[d]$. We clearly have $\sum_{j=m'}^{j'-1}P_{m',j|j'}=1$ and $\sum_{j=m}^{d}P_{m,j}=1$.

For each $m'\in [m]$ and $m'\le j\le d-m+m'$ let $z_{m',j}$, be the the sum of the products of numbers of $\mathcal{S}$ with indices $u_1, u_2, \ldots, u_{m'}$ contained in each $m'$-sized subset of $[j]$ such that $u_{m'}=j$. We define $Z_{m',k}:=\sum_{i=m'}^{k}z_{m',i}$ for any integer $k$ such that $m'\le k\le d-m+m'$. Thus, for all $m'\in [m-1]$ and $m'\le j< j'\le d-m+m'$ we have

\[
P_{m',j|j'}=\frac{z_{m',j}}{Z_{m',j'-1}}\,.
\]

Analogously, for the selection of $u_m$, for all $m\le j\le d$ we can write

\[
P_{m,j}=\frac{z_{m,j}}{Z_{m,d}}\,.
\]

Hence, once we obtain $z_{m',j}$ and $Z_{m',j'-1}$ for 
all $m'\in [m-1]$ and $m'\le j< j'\le d-m+m'$, $z_{m,j}$ for 
all $m\le j\le d$, and ${Z_{m,d}}$, we can immediately compute the desired probabilities to sample $u_m, u_{m-1},\ldots,u_1$ in this (reverse) order. 

We now show how to calculate these values. To this goal, since $Z_{m',k}:=\sum_{i=m'}^{k}z_{m',i}$, we only need to show how to compute the values appearing at the numerator in the above probability formulas. 

The possibility to efficiently the above probabilities is given by the following observation: 

\[
z_{m',j}=Z_{m'-1,j-1}\cdot q_j\,.
\]

Note that $Z_{m'-1,j-1}$ can be in turn defined in terms of 
$z_{m'-1,m'-1}$, $z_{m'-1,m'}$, $z_{m'-1,m'+1},\ldots, z_{m'-1,j-2}$, $z_{m'-1,j-1}$. This recurrence relation allows us to compute all these values once we know $z_{1,1}, z_{1,2},\ldots, z_{1,d}$. Since we clearly have $z_{1,j}=q_j$ for all $j\in [d]$, we can therefore compute all these values and the above probabilities to efficiently accomplish this sampling operation by finding the indices $u_m,u_{m-1},\ldots,u_1$ in this order. It is immediate to verify that both the number of sum and multiplication operations are equal to $\Theta(md)$.